\documentclass[journal]{IEEEtran}

\ifCLASSINFOpdf
\else
\fi
\usepackage{amsmath} 
\usepackage{verbatim} 
\usepackage{graphicx}
\usepackage[table,xcdraw]{xcolor}
\usepackage{tabularx}
\usepackage{verbatim}
\usepackage{subfig}
\usepackage{multirow}
\usepackage{float}
\usepackage{hyperref}

\AtBeginDocument{\usepackage{booktabs}}
\usepackage{color}

\usepackage{caption}
\usepackage{hhline}
\usepackage{float}

\usepackage{amssymb}
\usepackage{tabularx}
\newcolumntype{L}[1]{>{\raggedright\let\newline\\\arraybackslash\hspace{0pt}}m{#1}}
\newcolumntype{C}[1]{>{\centering\let\newline\\\arraybackslash\hspace{0pt}}m{#1}}
\newcolumntype{R}[1]{>{\raggedleft\let\newline\\\arraybackslash\hspace{0pt}}m{#1}}

\graphicspath{ {Images/}}

\begin{document}


\title{Multi-scale self-guided attention for medical image segmentation} 



\author{Ashish~Sinha
 and~Jose~Dolz
\IEEEcompsocitemizethanks{\IEEEcompsocthanksitem A. Sinha is with the Indian Institute of Technology Roorkee, India. e-mail: asinha@mt.iitr.ac.in.} 
\IEEEcompsocitemizethanks{\IEEEcompsocthanksitem J. Dolz is with the \'Ecole de technologie Superieure, Montreal, Canada. email:jose.dolz@etsmtl.ca.}
\thanks{Manuscript received XXX; revised XXX.}}


\maketitle
\begin{abstract}

Even though convolutional neural networks (CNNs) are driving progress in medical image segmentation, standard models still have some drawbacks. First, the use of multi-scale approaches, i.e., encoder-decoder architectures, leads to a redundant use of information, where similar low-level features are extracted multiple times at multiple scales. Second, long-range feature dependencies are not efficiently modeled, resulting in non-optimal discriminative feature representations associated with each semantic class. In this paper we attempt to overcome these limitations with the proposed architecture, by capturing richer contextual dependencies based on the use of guided self-attention mechanisms. This approach is able to integrate local features with their corresponding global dependencies, as well as highlight interdependent channel maps in an adaptive manner. Further, the additional loss between different modules guides the attention mechanisms to neglect irrelevant information and focus on more discriminant regions of the image by emphasizing relevant feature associations. We evaluate the proposed model in the context of semantic segmentation on three different datasets: abdominal organs, cardiovascular structures and brain tumors. A series of ablation experiments support the importance of these attention modules in the proposed architecture. In addition, compared to other state-of-the-art segmentation networks our model yields better segmentation performance, increasing the accuracy of the predictions while reducing the standard deviation. This demonstrates the efficiency of our approach to generate precise and reliable automatic segmentations of medical images. Our code is made publicly available at: \textcolor{black}{\href{https://github.com/sinAshish/Multi-Scale-Attention}{https://github.com/sinAshish/Multi-Scale-Attention}}


\end{abstract}



\maketitle 

\section{Introduction}


Semantic segmentation of medical images is a crucial step in diagnosis, treatment and follow-up of many diseases. Despite the automation of this task has been widely studied in the past, manual annotations are still typically used in clinical practice, which is a time-consuming and prone to inter and intra-observer variability process. Thus, there is a high demand on accurate and reliable automatic segmentation methods that allow to improve the work flow efficiency in clinical scenarios, alleviating the workload of radiologists and other medical experts.  



Recently, convolutional neural networks (CNNs) have achieved state-of-the-art performance in a breadth of visual recognition tasks, becoming very popular due to their powerful, nonlinear feature extraction capabilities. These deep models dominate the literature in medical image segmentation \cite{litjens2017survey} and have achieved outstanding performance in a broad span of applications, including brain \cite{dolz2018hyperdense} or cardiac \cite{bernard2018deep} imaging, for example, becoming the \textit{de facto} solution for these problems. In this scenario, fully convolutional neural networks \cite{FCN} or encoder-decoder architectures \cite{ronneberger2015u,lin2017refinenet} are typically the standard choice. These architectures are commonly composed of a contracting path, which collapses an input image into a set of high-level features, and an expanding path, where high-level features are used to reconstruct a pixel-wise segmentation mask at a single \cite{FCN} or multiple upsampling steps \cite{ronneberger2015u,lin2017refinenet}. Nevertheless, despite their strong representation power, these multi-scale approaches lead to a redundant use of information flow, e.g., similar low-level features are extracted multiple times at different levels within the network. Furthermore, the discriminative power of the learned feature representations for pixel-wise recognition may be insufficient for some challenging tasks, such as medical image segmentation.

Recent works to improve the discriminative ability of feature representations include the use of multi-scale context fusion \cite{chen2018deeplab,zhao2017pyramid,chen2018encoder,yu2015multi}. Zhao et al. \cite{zhao2017pyramid} proposed a pyramid network that exploited global information at different scales by aggregating feature maps generated by multiple dilated convolutional blocks. Aggregation of contextual multi-scale information can also be achieved through pooling operations \cite{liu2015parsenet}. Even though these strategies may help to capture objects at different scales, contextual dependencies for all image regions are homogeneous and non-adaptive, ignoring the difference between local representation and contextual dependencies for different categories. Further, these multi-context representations are manually designed, lacking flexibility to model the multi-context representations. This makes that long-range object relationships in the whole image cannot be fully leveraged in these approaches, which is of pivotal importance in many medical imaging segmentation problems.






Alternatively, attention mechanisms have been widely studied in deep CNNs for many computer vision tasks in order to efficiently integrate local and global features, including human pose estimation \cite{chu2017multi}, emotion recognition \cite{gupta2018attention}, text detection \cite{huang2019mask}, object detection \cite{chen2018reverse} and classification \cite{li2018tell}. Unlike standard multi-scale features fusion approaches, which compress an entire image into a static representation, attention allows the network to focus on the most relevant features without additional supervision, avoiding the use of multiple similar feature maps and highlighting salient features that are useful for a given task. Semantic segmentation networks have also benefited from attention modules, which has resulted in enhanced models for pixel-wise recognition tasks \cite{chen2016attention,zhao2018psanet,fu2018dual,li2018pyramid,yu2018bisenet,zhang2019deep}. For example, Chen et \textit{al.} \cite{chen2016attention} proposed an attention mechanism to weight multi-scale features extracted at different scales in the context of natural scene segmentation. This method improved the segmentation performance over classical average and max-pooling techniques to merge multi-scale features predictions.

Despite the growing interest on integrating attention mechanisms in image segmentation networks for natural scenes, their adoption in medical images remains scarce \cite{wang18d,li2018attention,schlemper2019attention,nie2018asdnet,roy2018concurrent}, being limited to simple attention models. Thus, in this work, we explore more complex attention mechanisms that can boost the performance of standard deep networks for the task of medical image segmentation. Specifically, we propose a multi-scale guided attention network for medical image segmentation. First, the multi-scale approach generates stacks at different resolutions containing different semantics. While lower-level stacks focus on local appearance, higher-level stacks will encode global representations. This multi-scale strategy encourages that attention maps generated at different resolutions encode different semantic information. Then, at each scale, a stack of attention modules will gradually remove noisy areas and emphasize those regions that are more relevant to the semantic descriptions of the targets. Each attention module contains two independent self-attention mechanisms, which focus on modelling position and channel feature dependencies, respectively. This duple allows to model wider and richer contextual representations and improve dependencies between channel maps, resulting in enhanced feature representations. We validate our method in \textcolor{black}{three different segmentation tasks: abdominal organ, cardiovascular structures and brain tumor}. Results show that the proposed architecture improves the segmentation performance by successfully modeling rich contextual dependencies over local features.

\section{Related work}

\subsection{Medical image segmentation}
Even though segmentation of medical images has been widely studied in the past \cite{heimann2009statistical, dolz2015segmentation} it is undeniable that CNNs are driving progress in this field, leading to outstanding performances in many applications. Most available medical image segmentation architectures are inspired from the well-known fully convolutional neural network (FCN) \cite{FCN} or UNet \cite{ronneberger2015u}. In FCN the fully connected layers of standard classification CNNS are replaced by convolutional layers to achieve dense pixel prediction at one forward step. To recover the original resolution of the input image, the prediction is upsampled in a single step. Further, to improve the prediction capabilities, skip connections are included in the network by employing the intermediate feature maps. On the other hand, UNet contains contractive and expansive paths created using the combination of convolutional layers with pooling and upsampling layers. Skip connections are used to concatenate the features from contractive and expansive path layers. Many extensions of these networks have been proposed to solve pixel-wise segmentation problems in a wide range of applications \cite{fechter2017esophagus,li2018h,man2019deep,dolz20183d,carass2018comparing,zotti2018convolutional,dolz2018multiregion,jin2019accurate,heinrich2019obelisk,wang2019abdominal}.

\subsection{Deep attention}

Attention mechanisms aim at emphasizing important local regions captured in local features and filtering irrelevant information transferred by global features, improving the modeling of long-range dependencies. These modules have therefore become an essential part of models that need to capture global dependencies. The integration of these attention modules has been proved very successful in many vision problems, such as image captioning \cite{pedersoli2017areas}, image question-answering \cite{yang2016stacked} or classification \cite{wang2017residual}. Self-attention \cite{parikh2016decomposable,vaswani2017attention,wang2018non} has recently attracted the attention of researchers, as it exhibits a good ability to model long-range dependencies while maintaining computational and statistical efficiency. In these modules, the response at each position is calculated by attending to all positions and taking their weighted average in an embedding space. For image vision problems, \cite{zhao2018psanet,fu2018dual} integrated self-attention to model the relation of local features with their corresponding global dependencies. For instance, the point-wise spatial attention network (PSANet) proposed in \cite{zhao2018psanet} allows a flexible and dynamic aggregation of long-range contextual information by connecting each position in the feature map with all the others through self-adaptive attention maps.


Recent works have indicated that attention features generated in a single step may still contain noise introduced from regions that are irrelevant for a given class, leading to sub-optimal results \cite{yang2016stacked,ji2018stacked}. To overcome this issue, some works have investigated the use of progressive multiple attention layers in the context of visual question answering \cite{yang2016stacked} or zero shot learning \cite{ji2018stacked}. This strategy gradually filters undesired noise and emphasizes the regions highly relevant for the class semantic representations. To the best of our knowledge, the application of stacked attention modules remains unexplored in semantic segmentation.

\subsection{Medical image segmentation with deep attention}

Even though attention mechanisms are becoming popular on many vision problems, the literature on medical image segmentation with attention remains scarce, with simple attention modules \cite{wang18d,li2018attention,schlemper2019attention,nie2018asdnet,roy2018concurrent}. Wang et \textit{al.} \cite{wang18d} employed attention modules at multiple resolutions to combine local deep attention features (DAF) with global context for prostate segmentation on Ultrasound images. To model long-range dependencies local and global features were combined in a simple attention module, which contains three convolutional layers followed by a softmax function to create the attention map. A similar attention module, composed of two convolutional layers followed by a softmax, was integrated in a hierarchical aggregation framework integrated in UNet for left atrial segmentation \cite{li2018attention}. More recently, additive attention gate modules were integrated in the skip connections of the decoding path of UNet with the goal of better model complimentary information from the encoder \cite{schlemper2019attention}.

\section{Methods}\label{sec:methods}

\subsection{Overview}

\begin{figure}
    \centering
    \includegraphics[width=.45\textwidth]{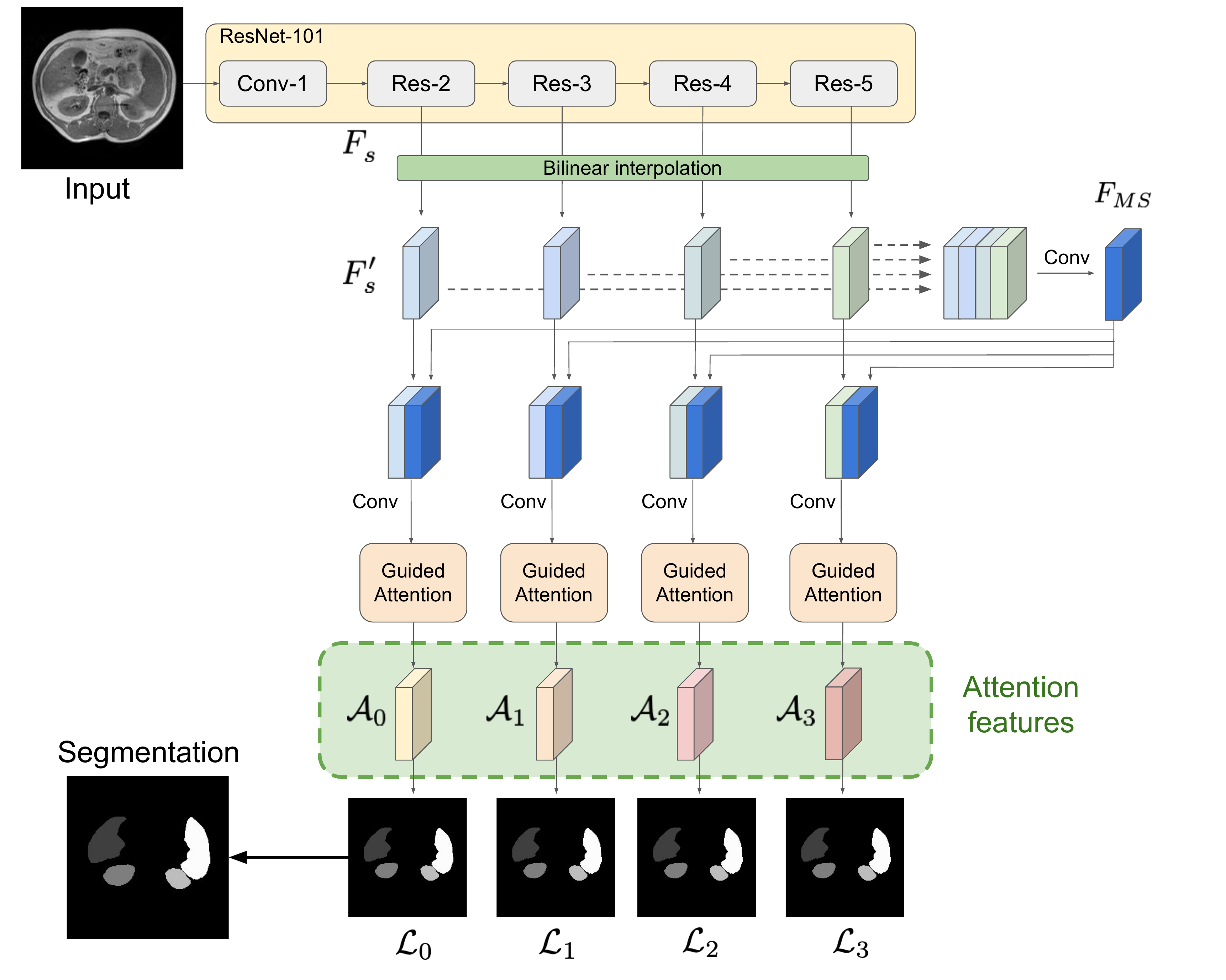}
    \caption{Overview of the proposed multi-scale guided attention network. We resort to ResNet-101 to extract dense local features. \textcolor{black}{ Four feature maps with different sizes --acquired from the outputs of [“Res-2”, “Res-3”, “Res-4”, “Res-5”]-- are employed. The guided attention modules will generate attentive features at multiple scales, removing noisy areas and helping the network to emphasize the regions that are more relevant to the semantic classes. }}
    \label{fig:overview}
\end{figure}

Target structures on medical imaging typically present intra and inter-class diversity on size, shape and texture, particularly if images are processed in 2D. Traditional CNNs for segmentation have a local receptive field, which results in the generation of local feature representations. As long-range contextual information is not properly encoded, local features representations may lead to potential differences between features corresponding to the pixels with the same label \cite{fu2018dual}. This may introduce intra-class inconsistency that can ultimately impact on the recognition performance \cite{peng2017large}. To tackle with this problem, we investigate attention mechanisms to build associations between features. First, global context is captured by employing a multi-scale strategy. Then, learned features at multiple scales are fed into the guided attention modules, which are composed by a stack of spatial and channel self-attention modules. While the spatial and channel self-attention modules will help to adaptively integrate local features with their global dependencies, the stack of attention modules will help to gradually filter noise out emphasizing on relevant information. The overview of the proposed framework is depicted in Figure \ref{fig:overview}.



\subsection{Multi-scale attention maps}

Multi-scale features are known to be useful in computer vision problems even before the deep learning era \cite{arbelaez2010contour}. In the context of deep segmentation networks, the integration of multi-scale features has demonstrated astonishing performance \cite{chen2016attention,hariharan2015hypercolumns,mostajabi2015feedforward}.
Inspired by these works we make use of learned features at multiple scales, which help to encode both global and local context. Specifically we follow the multi-scale strategy recently proposed in \cite{wang18d}. In this setting, features at multiple scales are denoted as $F_s$, where $s$ indicates the level in the architecture (Fig. \ref{fig:overview}). Since features come at different resolutions for each level $s$, they are upsampled to a common resolution by employing \textcolor{black}{bilinear} interpolation, leading to enlarged feature maps $F'_s$. Then, $F'_s$ from all the scales are concatenated forming a tensor that is convolved to create a common multi-scale feature map, $F_{MS}=conv([F'_0,F'_1,F'_2,F'_3])$. \textcolor{black}{Thus, $F_{MS}$ encodes low-level detail information from shallow layers as well as high-level semantics learned in deeper layers.} Then, this new multi-scale feature map is combined with each of the feature maps at different scales $s$ and fed into the guided attention modules to generate the attention features $\mathcal{A}_s$:

\begin{equation}
    \mathcal{A}_{s} = AttMod_{s}(conv([F'_{s},F_{MS}]))
\end{equation}

where $AttMod$ represents each guided attention module \textcolor{black}{(Section \ref{ssec:guide})}. \textcolor{black}{As the multi-scale feature maps $F_{MS}$ are combined at each individual layer, complementary low-level information and high-level semantics from $F_{MS}$ are encoded jointly, resulting in a more powerful representation. In the following sections we detail how the attentive features A$_s$ are obtained.}

\subsection{Spatial and Channel self-attention modules}
\begin{figure}[h!]
    \centering
    \includegraphics[width=.475\textwidth]{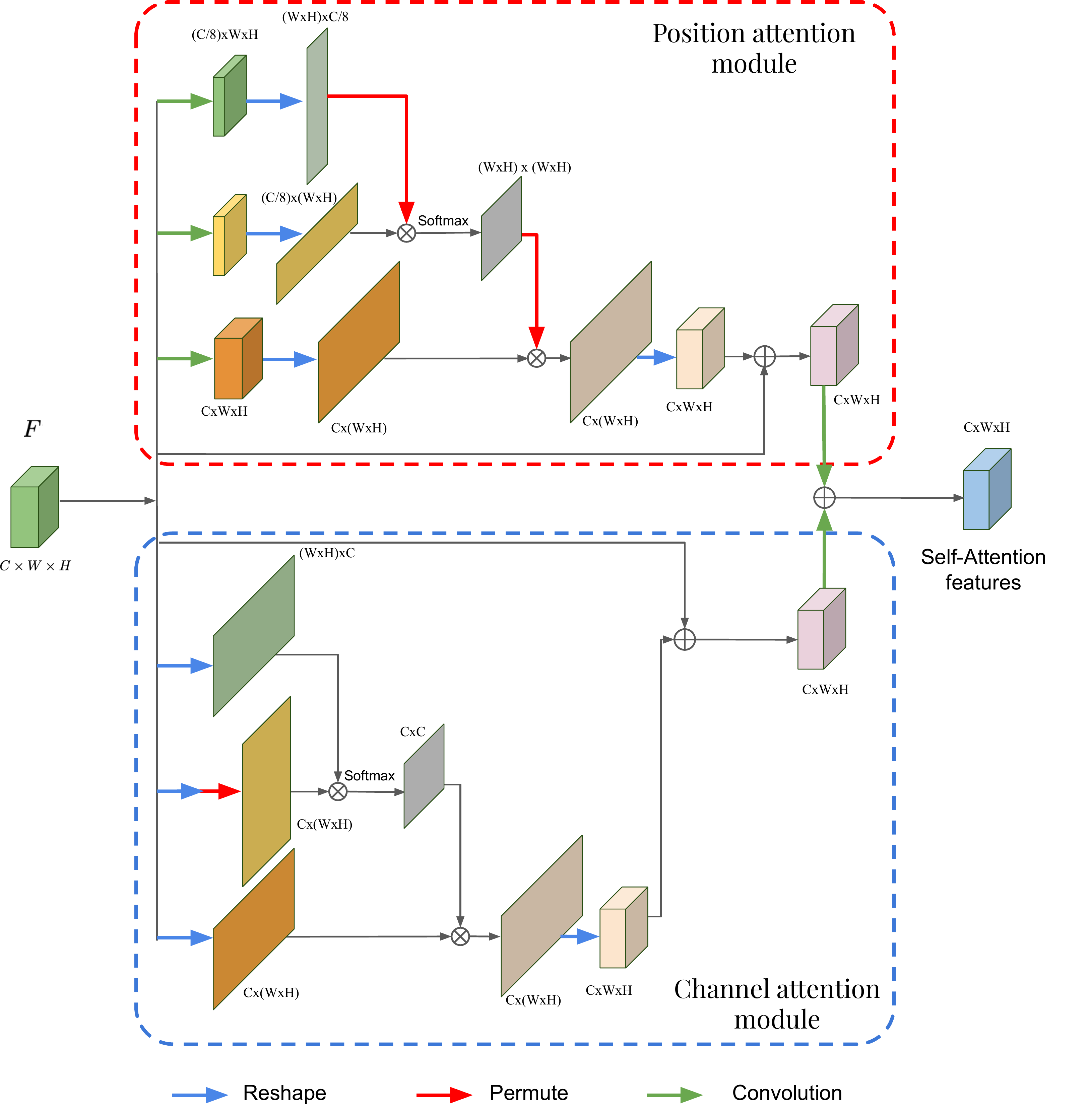}
    \caption{Details of the position and channel attention modules inspired by \cite{fu2018dual}.}
    \label{fig:dual}
\end{figure}

As introduced earlier, receptive fields in traditional segmentation deep models are reduced to a local vicinity. This limits the capabilities of modeling wider and richer contextual representations. On the other hand, channel maps can be considered as class-specific responses, where different semantic responses are associated with each other. Thus, another strategy to enhance the feature representation of specific semantics is to improve the dependencies between channel maps \cite{chen2017sca}. To address these limitations of standard CNNs we employ the position and channel attention modules recently proposed in \cite{fu2018dual}, which are depicted in Figure \ref{fig:dual}.

\paragraph*{\textbf{Position attention module (PAM)}}Let denote $F \in \mathbb{R}^{C\times W\times H}$ an input feature map to the attention module, where $C,W,H$ represent the channel, width and height dimensions, respectively. In the upper branch $F$ is passed through a convolutional block, resulting in a feature map $F_{0}^p \in \mathbb{R}^{C'\times W\times H}$, where $C'$ is equal to $C/8$\footnote{We use the superscript $p$ to indicate that the feature map belongs to the position attention module. Similarly, we will employ the superscript $c$ for the channel attention module features.}. Then, $F_{0}^p$ is reshaped to a feature map of shape $(W\times H)\times C'$. In the second branch, the input feature map $F$ follows the same operations and then is transposed, resulting in $F_{1}^p \in \mathbb{R}^{C'\times (W\times H)}$. Both maps are multiplied and softmax is applied on the resulted matrix to generate the spatial attention map $S^p \in \mathbb{R}^{(W \times H) \times (W \times H)}$:

\begin{equation}
    s_{i,j}^p = \frac{\exp{(F_{0,i}^p \cdot F_{1,j}^p)}}{\sum_{i=1}^{W \times H}\exp{(F_{0,i}^p \cdot F_{1,j}^p)}}
\end{equation}

where $s_{i,j}^p$ evaluates the impact of the $i^{th}$ position on the $j^{th}$ position. The input $F$ is fed into a different convolutional block in the third branch, resulting in $F_{2}^p \in \mathbb{R}^{C\times (W\times H)}$, which has the same shape as $F$. As in the other branches, $F_{2}^p$ is reshaped becoming $F_{2}^p \in \mathbb{R}^{C\times (W\times H)}$. Then it is multiplied by a permuted version of the spatial attention map $S$, whose output is reshaped to a $\mathbb{R}^{C\times (W\times H)}$. The attention feature map corresponding to the position attention module, i.e., $F_{PAM}$, can be therefore formulated as follows:

\begin{equation}
    F_{PAM,j} = \lambda_p\sum_{i=1}^{W\times H}s_{ij}^pF_{2,j}^p + F_j
\end{equation}

As in \cite{fu2018dual}, the value of $\lambda_p$ is initialized to 0 and it is gradually learned to give more importance to the spatial attention map. Thus, the position attention module selectively aggregates global context to the learned features, guided by the spatial attention map. 

\paragraph*{\textbf{Channel attention module (CAM)}}
The pipeline of the channel attention module is depicted at the bottom of Figure \ref{fig:dual}. The input $F \in \mathbb{R}^{C\times W\times H}$ is reshaped in the first two branches of the CAM, and permuted in the second branch, leading to $F_{0}^c \in \mathbb{R}^{(W\times H)\times C}$ and 
$F_{1}^c \in \mathbb{R}^{C \times (W\times H)}$, respectively. Then, we perform a matrix multiplication between $F_{0}^c$ and $F_{1}^c$, and obtain the channel attention map $S^c \in \mathbb{R}^{C \times C}$ as: 

\begin{equation}
    s_{i,j}^c = \frac{\exp{(F_{0,i}^c \cdot F_{1,j}^c)}}{\sum_{i=1}^{C}\exp{(F_{0,i}^c \cdot F_{1,j}^c)}}
\end{equation}

where the impact of the $i^{th}$ channel on the $j^{th}$ is given by $s_{i,j}^c$. This is then multiplied by a transposed version of the input $F$, i.e., $F_2^c$, whose result is reshaped to $\mathbb{R}^{C \times (W\times H)}$. Then, the final channel attention map is obtained as:

\begin{equation}
    F_{CAM,j} = \lambda_c\sum_{i=1}^{C}s_{ij}^cF_{2,j}^c + F_j
\end{equation}

where $\lambda_c$ controls the importance of the channel attention map over the input feature map $F$. Similarly to $\lambda_p$, $\lambda_c$ is initially set to 0 and gradually learned. This formulation aggregates weighted versions of the features of all the channels into the original features, highlighting class-dependent feature maps and increasing feature discriminability between classes. At the end of both attention modules, the new generated features are fed into a convolutional layer before performing an element-wise sum operation to generate the position-channel attention features.

\subsection{Guiding attention}
\label{ssec:guide}
\begin{figure}[h!]
    \centering
    \includegraphics[width=.45\textwidth]{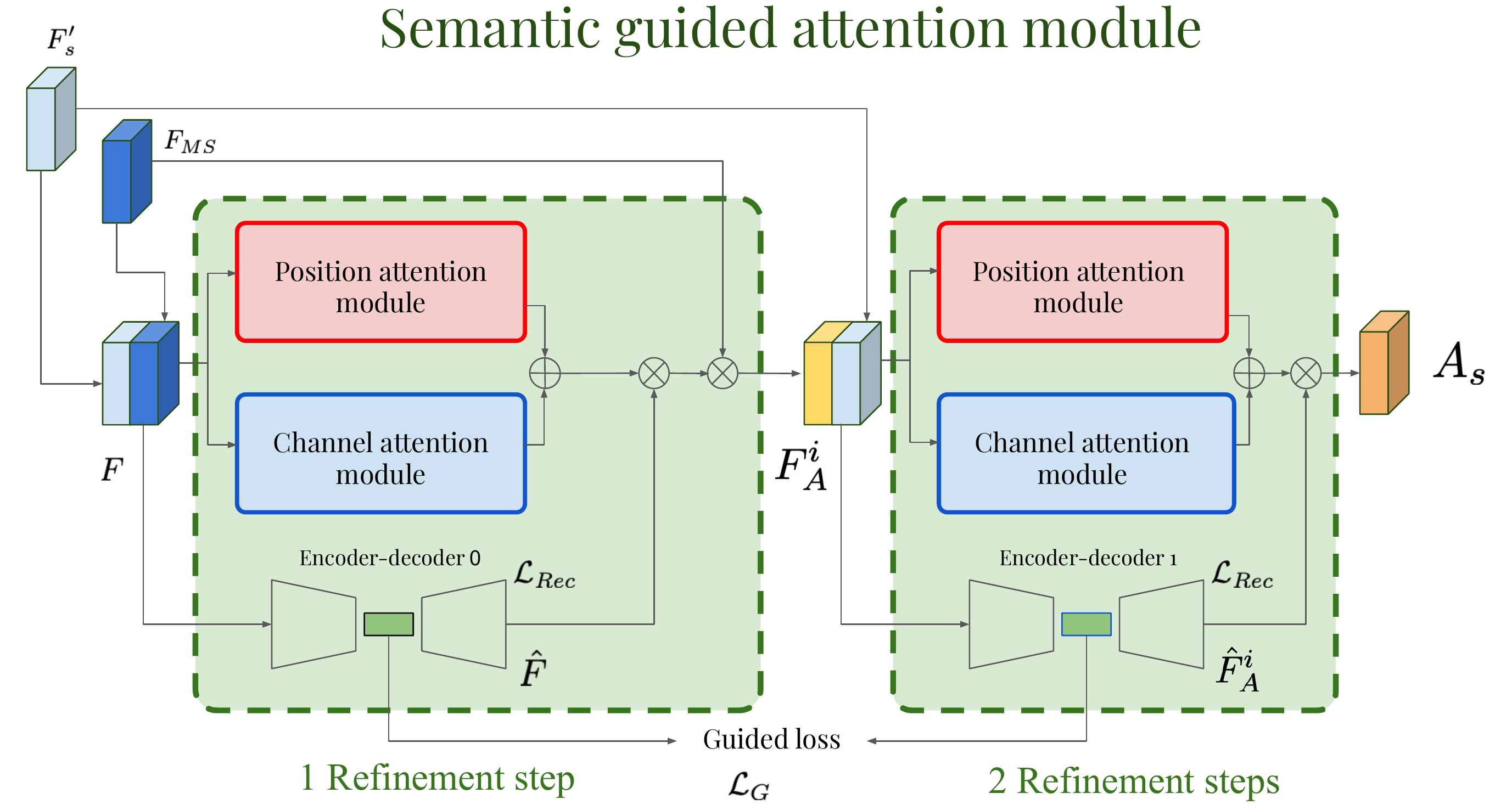}
    \caption{An illustration of the semantic guided attention module \textcolor{black}{with 2 refinement steps. For each scale $s$ this module provides a set of attentive features, i.e., A$_s$.}}
    \label{fig:guidedModule}
\end{figure}

\textcolor{black}{Inspired by recent work to stack attention modules in the context of image classification \cite{ji2018stacked}, we propose to add progressive refinement of attentive features through sequential refinement modules. The intuition is that this sequential refinement will progressively weight the importance of different local regions, while masking out irrelevant noise. Particularly,} given the feature map $F$ at the input of the guided attention module at scale $s$--generated by concatenating $F_{MS}$ and $F'_{s}$--, it generates attention features via a multi-step refinement \textcolor{black}{(Fig. \ref{fig:guidedModule})}. In the first step, $F$ is used by the position and channel attention modules to generate self-attention features. In parallel, we integrate an encoder-decoder network that compresses the input features $F$ into a compacted representation in the latent space \textcolor{black}{\cite{ji2018stacked}}. The objective is that the class information can be embedded \textcolor{black}{into the subsequent guided attention modules} by forcing the \textcolor{black}{latent} representation of encoder-decoders to be close, which is formulated as:


\begin{equation}
    \textcolor{black}{\mathcal{L}_{G} = \sum_{i}^{M-1} \|\mathbb{E}_i(F_{A}^{i-1})-\mathbb{E}_{i+1}(F_{A}^i)\|_2^2}
\end{equation}

where \textcolor{black}{$\mathbb{E}_i(\cdot)$ is the encoded representation of the \textit{i}-th encoder-decoder network, $F_{A}^i$ denotes the attention features generated after the \textit{i}-th dual attention module and $M$ the number of iterations. Note that $F_{A}^{i-1}$ are the features at the input of the semantic guided attention module, $F$.} Specifically, the feature maps reconstructed in the first encoder-decoder ($n=0$) are combined with the self-attention features generated by the first attention module through a matrix-multiplication to generate $F_{SA}$. In addition, to ensure that the reconstructed features correspond to the features at the input of the position-channel attention modules, the output of the encoders are forced to be close to their input: 


\begin{equation}
    \textcolor{black}{\mathcal{L}_{Rec} = \sum_i^{M}\|F_i-\hat F_i\|_2^2}
    \label{eq:rec_loss}
\end{equation}

\textcolor{black}{where $\hat F_i$ are the reconstructed feature maps, i.e., $\mathbb{D}_i(\mathbb{E}_i(F))$ of the $i$-th encoder-decoder networks. }

As the guided attention module is applied at multiple scales, the combined guided loss for all the modules will be:

\begin{equation}
    \mathcal{L}_{G_{Total}} = \sum_{s=0}^S\mathcal{L}_{G}^s
\end{equation}

Similarly, the total reconstruction loss becomes:

\begin{equation}
    \mathcal{L}_{Rec_{Total}} = \sum_{s=0}^S\mathcal{L}_{Rec}^s
\end{equation}

where $\mathcal{L}_{Rec_1}$ and \textcolor{black}{$\mathcal{L}_{Rec_2}$} are the reconstruction losses for the encoder-decoder architectures in the first and second block of the guided attention module.

\subsection{Deep supervision}

While the attention modules do not require auxiliary objective functions, we found that the use of extra supervision at each scale \cite{lee2015deeply} \textcolor{black}{improved the segmentation performance of the proposed model}, which is in line with similar works in the literature \cite{chen2016attention,wang18d,schlemper2019attention}. 

\begin{equation}
   \mathcal{L}_{Seg_{Total}} = \sum_{s=0}^S\mathcal{L}_{Seg_{F'}}^s + \sum_{s=0}^S\mathcal{L}_{Se\textbf{\textbf{}}g_A}^s 
   \label{eq:seg_loss}
\end{equation}

where the first term refers to the segmentation results at the raw features $F'_s$ and the second term evaluates the segmentation result provided by the attention features. In all the cases, the multi-class cross-entropy between the network prediction and the ground truth labels is employed as segmentation loss. The final objective function to optimize becomes:


\begin{equation}
    \mathcal{L}_{Total} = \alpha \mathcal{L}_{Seg_{Total}} + \beta \mathcal{L}_{G_{Total}}
    + \gamma \mathcal{L}_{Rec_{Total}}
    \label{eq:total_loss}
\end{equation}

where $\alpha$, $\beta$ and $\gamma$ control the importance of each term in the main loss function.

\section{Experiments}
\label{sec:Experiments}


\subsection{\textbf{Experimental setting}}

\paragraph*{\textbf{Datasets}}We employ \textcolor{black}{three} public segmentation benchmarks to evaluate our method. First, the abdominal MRI dataset from the Combined Healthy Abdominal Organ Segmentation (CHAOS) Challenge \cite{selver2014exploring,selvi2015segmentation,selver2014segmentation}. Particularly, we focus on the segmentation of abdominal organs \textcolor{black}{(spleen, liver and kidneys)} on MRI (T1-DUAL in phase), which includes scans from 20 subjects for training with their corresponding ground truth annotations, and 20 for testing without annotations. Scans have a resolution of 256$\times$256 pixels per slice, and between 26 and 50 slices. Since testing labels are not provided within the dataset, we employed the training dataset for our experiments, splitting it into subsets of 13, 2 and 5 subjects that were used for training, validation and testing. We repeated the process 3 times selecting different subjects and report the average results over the three folds. Then, we evaluated our approach on the task of whole-heart and great vessel segmentation from 3D Cardiovascular MRI in congenital heart disease, provided in the HVSMR 2016 Challenge \cite{pace2015interactive}. \textcolor{black}{Particularly, the myocardium and the blood pool are targeted in this scenario}. The training set consists on 10 MRI Axial scans with their corresponding manual segmentations. Image dimensions varied across subjects, with an average of 390 $\times$ 390 $\times$ 165 voxel volumes. We report results on the training data, by employing a 5-fold cross-validation strategy, where each fold contains 6 scans for training, 2 for validation and 2 for testing. To increase the variability of the data, we rotate, flipped and mirrored the images randomly, but without augmenting the dataset size. \textcolor{black}{For the third task, we employed the brain segmentation dataset provided in the Medical Segmentation Decathlon Challenge\footnote{http://medicaldecathlon.com}. Particularly, this dataset contains multimodal multisite MRI data (FLAIR, T1w, T1gd,T2w) from the BRATS'16 and BRATS'17 Challenges \cite{menze2014multimodal,bakas2017advancing,bakas2018identifying}. The focus of this dataset is on the segmentation of necrotic (TC) and active areas (ET), as well as oedema (ED) in brain gliomas. We employed 484 scans that were split into training (388 scans), validation (48 scans) and testing (48 scans). Similarly to previous tasks, we rotate, flipped and mirrored the images randomly, but without augmenting the dataset size.}

\paragraph*{\textbf{Network architectures}}
The multi-scale strategy in the proposed network is based on the recently work in \cite{wang18d}, and is considered as the lower baseline in our experiments. First, we perform an ablation study on the different proposed modules to evaluate the impact of each choice in the segmentation performance. The first two networks --i.e., \textit{Proposed (PAM)} and \textit{Proposed (CAM)}-- extend the baseline by replacing the attention module by either the spatial or the channel self-attention module (Fig. \ref{fig:dual}), respectively. Then, both modules are combined simultaneously, leading to the \textit{Proposed (DANet)} model. In the next model --i.e., \textit{Proposed (MS-Dual)}-- the attention features generated by the dual attention module are refined in a multi-step process, where a second dual attention module is included. Last, the proposed architecture, referred to as \textit{Proposed (MS-Dual-Guided)} extends the \textit{Proposed (MS-Dual)} model by incorporating the semantic guidance (Fig. \ref{fig:guidedModule}). \textcolor{black}{We also evaluated the impact of different elements, other than the attention modules, on the proposed multi-scale architectures. First, we remove the deep supervision (first term in eq. \ref{eq:seg_loss}) on our model. Second, instead of using an encoder-decoder structure to reconstruct the input features at each dual attention module, we remove this and replace the eq. \ref{eq:rec_loss} by the mean error square loss between the input and the output of each attention module. This models is referred to as \textit{w/out encoder-decoder (dist)}. And last, we also investigated the effect of not having an encoder-decoder, i.e., no guidance, in the refinements steps, which is referred to as \textit{w/out encoder-decoder)}. In addition, we evaluated the impact of having multiple refinements steps $n$, with $n=1,2,3$ and $5$. }

Furthermore we compared the performance of the proposed network to other state-of-the-art architectures integrating attention: Attention UNet \cite{schlemper2019attention}, DANet \cite{fu2018dual} and Pyramidal Attention Network (PAN) \cite{li2018pyramid}. 

\paragraph*{\textbf{Training and implementation details}}

We train all the networks using Adam optimizer with mini-batch of size 8, and with $\beta_1$ and $\beta_2$ set to 0.9 and 0.99, respectively. While most of the networks converged during the first 250 epochs, we found that PAN \cite{li2018pyramid} and DANet \cite{fu2018dual} needed around 400 epochs to achieve the best results. The learning rate is initially set to 0.001 and multiplied by 0.5 after 50 epochs without improvement on the validation set. As a segmentation objective function, we employ the cross-entropy error at each pixel over all the categories for all the networks. Furthermore, as introduced in Section \ref{sec:methods}, we use the objective function in eq. (\ref{eq:total_loss}) in the proposed architecture, with $\alpha$, $\beta$ and $\gamma$ set empirically to 1, 0.25 and 0.1, respectively. As input of the networks we employed 2D axial images of size 256 $\times$ 256. Experiments were performed in a server equipped with a Titan V. The code of our model is made publicly available at https://github.com/sinAshish/Multi-Scale-Attention.


\paragraph*{\textbf{Evaluation}}

Similarity between ground truth and CNN segmentations is assessed by employing several comparison metrics. First, we resort to the widely used Dice similarity coefficient (DSC) to compare volumes based on their overlap. Further, we also assess the segmentation performance based on the volume similarity (VS). Additionally, to measure the sensitivity to segmentation outline, we considered the use of the mean surface distance (MSD). 
The formulation of these metrics is detailed in the Supplemental materials. Since inter-slice distances and x-y spacing for each individual scan are not provided, we report these results on voxels.

\subsection{\textbf{Results}}

\paragraph*{\textbf{Ablation study on the proposed attention modules}}

To validate the individual contribution of different components to the segmentation performance, we perform an ablation experiment under different settings. Table \ref{table:metrics_ablation} reports the results of the different attention modules. Compared to the baseline, we observe that by integrating either a spatial (PAM) or an attention module (CAM) at each scale in the baseline architecture the performance improves between 2-3\% in terms of overlapping and volume similarity, and between 12-18\% in terms of surface distances, as average. On the other hand, having both modules in parallel --i.e., \textit{Proposed (DANet)}-- brings slightly better results in terms of DSC, but achieves lower performance when employing the surface distance metric. However, despite the lower average performance on the MSD, the proposed DANet model still achieves better results in 3 out of 4 structures compared to the channel attention model. This trend is repeated on the DSC metric, where DANet surpasses the proposed CAM architecture in the same 3 structures: liver and both left and right kidneys. This suggests that, even though both spatial and channel attention bring an improvement on the performance, the channel attention module contributes more than the spatial attention when they are combined. If features generated by the proposed DANet model are refined in a second step --network referred to as \textit{Proposed(MS-Dual)}-- the average results are further improved by nearly 0.7\% and 10\% in volume and distance-based metrics, respectively. Last, the introduction of the semantic-guided loss --\textit{Proposed (MS-Dual-Guided)}-- results in an additional boost on performance, yielding to the best values in the three metrics: 86.75\%(DSC), 93.85\%(VS) and 0.66 voxels (MSD). These results represent an improvement of 4.5\%, 4\% and 26\% in DSC, VS and MSD, respectively, compared to the baseline in \cite{wang18d}, showing the efficiency of the proposed attention network compared to individual attention components.

\begin{table}[ht!]
\centering
\scriptsize
\begin{tabular}{lcccc|c}\\
\toprule
Method & \textbf{DSC (\%)} & \textbf{VS (\%)} & \textbf{MSD (voxels)}  \\
\midrule
Baseline (DAF \cite{wang18d})  &  82.48 ($\pm$6.06) &   89.68 ($\pm$4.48) &  0.92 ($\pm$0.33)
\\
Proposed (PAM)  & 84.46 ($\pm$6.68) &  91.84 ($\pm$4.77) &  0.80 ($\pm$0.43) \\
Proposed (CAM)  & 85.08 ($\pm$5.62) & 92.18 ($\pm$5.07) & 0.74 ($\pm$0.32)\\
Proposed (DANet) &  85.52 ($\pm$5.86) &  92.07 ($\pm$5.23) &   0.77 ($\pm$0.41)      \\
Proposed (MS-Dual) &  \textcolor{blue}{\textbf{86.17 ($\pm$5.78)}} & \textcolor{blue}{\textbf{92.74 ($\pm$4.76)}} &  \textcolor{blue}{\textbf{0.67 ($\pm$0.30)}}\\
Proposed (MS-Dual-Guided)  &  \textcolor{red}{\textbf{86.75 ($\pm$5.05)}} & \textcolor{red}{\textbf{93.85 ($\pm$3.50)}} &  \textcolor{red}{\textbf{0.66 ($\pm$0.27)}}\\

\midrule
\end{tabular}
\caption{Ablation study on different attention modules on the Chaos dataset. The values show the average result of the experiments averaged over the 3 folds. Best and second best results are represented in red and blue bold, respectively.}
\label{table:metrics_ablation}
\end{table}

\begin{table}[ht!]
\centering
\scriptsize
\begin{tabular}{lcccc|c}\\
\toprule
\multicolumn{4}{c}{Proposed (MS-Dual and MS-Dual-Guided)} \\
\midrule
Model & \textbf{DSC (\%)} & \textbf{VS (\%)} & \textbf{MSD (voxels)}  \\
\midrule
\multicolumn{4}{l}{\textit{1 Refinement step}} \\
MS-Dual (No guidance)  & 85.75 ($\pm$5.08)& 92.72 ($\pm$3.65) & 0.71 ($\pm$0.28)\\
MS-Dual-Guided   & 86.34 ($\pm$5.17) & 93.47 ($\pm$3.78) & 0.68 ($\pm$0.29) \\
w/out deep supervision    & 84.71 ($\pm$4.86)  & 91.39 ($\pm$3.55) & 0.75 ($\pm$0.17) \\
w/out encoder-decoder (dist) &  85.92 ($\pm$5.17)  & 92.94 ($\pm$4.04) & 0.76 ($\pm$0.34)\\

\midrule
\multicolumn{4}{l}{\textit{2 Refinement steps}} \\
MS-Dual (No guidance)   & 86.17 ($\pm$5.78) & 92.74 ($\pm$4.76) &  \textcolor{blue}{\textbf{0.67 ($\pm$0.30)}}\\
MS-Dual-Guided   &  \textcolor{red}{\textbf{86.75 ($\pm$5.05)}} & \textcolor{red}{\textbf{93.85 ($\pm$3.50)}} &  \textcolor{red}{\textbf{0.66 ($\pm$0.27)}}\\
w/out deep supervision   & 83.51 ($\pm$5.52) &  91.80 ($\pm$3.66)& 0.75 ($\pm$0.16)\\
w/out encoder-decoder (dist)   &  \textcolor{blue}{\textbf{86.67 ($\pm$4.98)}}  & 93.67 ($\pm$3.38) & 0.77 ($\pm$0.31) \\

\midrule
\multicolumn{4}{l}{\textit{3 Refinement steps}} \\
MS-Dual (No guidance)  &  86.26 ($\pm$5.71) & 93.62 ($\pm$4.72) & 0.71 ($\pm$0.34)\\
MS-Dual-Guided   & 86.14 ($\pm$5.89)  & 93.50 ($\pm$3.98)& \textcolor{blue}{\textbf{0.67 ($\pm$0.36)}} \\
w/out deep supervision   & 83.22 ($\pm$5.72)& 90.95 ($\pm$4.31)& 0.80 ($\pm$0.17)\\
w/out encoder-decoder (dist)  &  85.88 ($\pm$4.78) & 93.23 ($\pm$3.71) &  0.79 ($\pm$0.39)\\
\midrule
\multicolumn{4}{l}{\textit{5 Refinement steps}} \\
MS-Dual (No guidance)  & 86.33 ($\pm$4.98)  &  \textcolor{blue}{\textbf{93.74 ($\pm$3.91)}} & 0.71 ($\pm$0.31)\\
MS-Dual-Guided   & 86.30 ($\pm$5.05)  & 93.16 ($\pm$4.11)& 0.68 ($\pm$0.22) \\
w/out deep supervision   & 83.88 ($\pm$5.78) & 91.03 ($\pm$3.66) & 0.87 ($\pm$0.34)\\
w/out encoder-decoder (dist)  & 86.16 ($\pm$4.23) &  92.98 ($\pm$2.93) & 0.80 ($\pm$0.31) \\
\midrule
\end{tabular}
\caption{\textcolor{black}{Ablation study on different elements on the MS-Dual and MS-Dual-Guided architectures evaluated on the Chaos dataset. The values show the average result of the experiments on the 3 folds. Best results are represented in red bold, while blue is used to highlight the second best performance.}}
\label{table:metrics_ablation2}
\end{table}

\textcolor{black}{The impact of the refinement steps, as well as of the several elements on both MS-Dual and MS-Dual-Guided models is reported in Table \ref{table:metrics_ablation2}. First, we can observe that increasing the number of refinement steps does not typically improve the performance of the methods. Indeed, best results are often obtained with only two attention guided modules. We argue that progressively refining feature maps may produce an excessive focus to the attentive regions, leading to strongly mined attentive features. This has an adverse effect, as the attentive features may concentrate in the most discriminative areas, not covering the whole extent of the object. Further, we observe that the proposed model including guided-attention outperforms all the variants, particularly in the distance-based metric. In addition, we provide a comparison in terms of complexity, whose results are depicted in Table \ref{tab:complexity}, in Supplemental Materials.}

\paragraph*{\textbf{Comparison to state-of-the-art}}

The experimental results obtained by several state-of-the-art segmentation networks are reported in Table \ref{table:sota_comp}. In the first dataset \textcolor{black}{(\textit{top})}, compared to other networks that were proposed in the context of medical image segmentation --i.e.,UNet \cite{ronneberger2015u}, Attention UNet \cite{schlemper2019attention} and DAF \cite{wang18d}-- our network achieves a mean improvement of 5.6\%, 4.3\% and 2.0\% (in terms of DSC), 4.9\%, 4.2\% and 2.1\% (on VS) and 25\%, 26\% and 6\% (on MSD), respectively. This difference in performance could be explained by the fact that the attention modules integrated in \cite{wang18d} and \cite{schlemper2019attention} are much simpler than those proposed in our architecture. On the other hand, attention modules on general computer vision tasks have attracted more attention, resulting in more elaborated strategies which typically achieve better segmentation results. Among these architectures, the PAN model \cite{li2018pyramid} with ResNet101 as backbone --the same as ours-- achieved the best results for segmentation networks proposed for natural scenes. Despite these competitive results, the proposed model still outperforms the PAN architecture by 2.4\%, 1.9\% and 12\% in DSC, VS and MSD. As PAN \cite{li2018pyramid} also employed a multi-scale architecture, these differences suggest that the use of dual self-attention and a guided refinement module can actually improve the performance of segmentation networks. Similarly, the proposed model outperforms other networks in the second \textcolor{black}{and third datasets \textcolor{black}{(\textit{middle and bottom})}}, indicating that it can be broadly applied to segmentation of medical images in general. Individual per-class scores for both datasets are given in Tables \ref{table:sota_comp_supplemental}, \ref{table:sota_comp_supplemental_hsvm} \textcolor{black}{and \ref{table:sota_comp_supplemental_brats}} in Supplemental Material. In addition to these values, we also depict the distribution of DSC, VS and MSD values on the 15 subjects used for evaluation in CHAOS for all the models (Fig. \ref{fig:metrics} in Supplemental Material).

\begin{table}[t!]
\centering
\scriptsize
\begin{tabular}{lccc}\\
\toprule
\multicolumn{4}{c}{\textbf{CHAOS}}\\
\midrule
Model & \textbf{DSC} & \textbf{VS} & \textbf{MSD} \\
 \midrule
UNet \cite{ronneberger2015u}  &  81.14 ($\pm$7.88) & 89.01 ($\pm$4.82) & 0.91 ($\pm$0.49)  \\
DANet \cite{fu2018dual}  &   83.89 ($\pm$9.54) & 91.42  ($\pm$4.52) &  0.78  ($\pm$0.23) \\
PAN (ResNet34) \cite{li2018pyramid}    & 82.70 ($\pm$6.51) & 90.32 ($\pm$5.27) & 0.86 ($\pm$0.29) \\
PAN (ResNet101)\cite{li2018pyramid}   & 84.34 ($\pm$6.17) & \textcolor{blue}{\textbf{91.93 ($\pm$4.71)}} & 0.78 ($\pm$0.31)  \\
DAF \cite{wang18d}  &   82.48 ($\pm$6.06)  & 89.68 ($\pm$4.48) &  0.92 ($\pm$0.33) \\
UNet Attention \cite{schlemper2019attention}  &  \textcolor{blue}{\textbf{84.77 ($\pm$5.27)}} & 91.79 ($\pm$3.53) & \textcolor{blue}{\textbf{0.72 ($\pm$0.24)}} \\
Proposed (MS-Dual-Guided)  &  \textcolor{red}{\textbf{86.75 ($\pm$5.05)}} & \textcolor{red}{\textbf{93.85 ($\pm$3.50)}} & \textcolor{red}{\textbf{0.66 ($\pm$0.27)}}  \\
\midrule
\multicolumn{4}{c}{\textbf{HSVM}}\\
\midrule
 & \textbf{DSC} & \textbf{VS} & \textbf{MSD}\\
 \midrule
UNet \cite{ronneberger2015u}  &  79.80 ($\pm$6.72) & 93.41 ($\pm$6.44) & 1.68 ($\pm$1.28)\\
DANet \cite{fu2018dual}  &   \textcolor{blue}{\textbf{82.55 ($\pm$5.91)}} & \textcolor{red}{\textbf{94.65 ($\pm$4.45)}} & 1.27 ($\pm$0.46) \\
PAN (ResNet34) \cite{li2018pyramid}    &  80.97 ($\pm$7.76) & 93.76 ($\pm$5.85) & 1.62 ($\pm$1.19) \\
PAN (ResNet101)\cite{li2018pyramid}   &  82.26 ($\pm$5.08) & 94.33 ($\pm$3.69) & 1.24 ($\pm$0.38) \\
DAF \cite{wang18d}  &    81.78 ($\pm$5.71)& 94.31 ($\pm$3.21) & 1.48 ($\pm$0.50) \\
UNet Attention \cite{schlemper2019attention}  & 81.58 ($\pm$6.84) & \textcolor{blue}{\textbf{94.61 ($\pm$4.17)}} & \textcolor{blue}{\textbf{1.25 ($\pm$0.42)}} \\
Proposed (MS-Dual-Guided)  &  \textcolor{red}{\textbf{83.20 ($\pm$4.93)}} & 94.45 ($\pm$2.39) & \textcolor{red}{\textbf{1.19 ($\pm$0.37)}} \\
\midrule
\multicolumn{4}{c}{\textbf{BRATS'18}}\\
\midrule
 & \textbf{DSC} & \textbf{VS} & \textbf{MSD}\\
 \midrule
UNet \cite{ronneberger2015u}  & 73.65 ($\pm$12.39) & 87.72 ($\pm$8.70) & 1.65 ($\pm$0.57)\\
DANet \cite{fu2018dual}  &  \textcolor{blue}{\textbf{79.09 ($\pm$10.89)}} & \textcolor{red}{\textbf{93.32 ($\pm$6.99)}} &  \textcolor{blue}{\textbf{0.95 ($\pm$0.33)}} \\
PAN (ResNet34) \cite{li2018pyramid}    &  74.12 ($\pm$12.76) & 89.85 ($\pm$9.93) &  1.42 ($\pm$0.52) \\
PAN (ResNet101)\cite{li2018pyramid}   &  76.89 ($\pm$11.53)  & 91.76 ($\pm$8.11) & 1.17 ($\pm$0.47) \\
DAF \cite{wang18d}  &  76.78 ($\pm$11.77) & 90.58 ($\pm$9.03) &  1.21 ($\pm$0.46) \\
UNet Attention \cite{schlemper2019attention} &   78.61 ($\pm$10.58) & 92.66 ($\pm$6.86) & 1.02 ($\pm$0.40) \\
Proposed (MS-Dual-Guided)  &   \textcolor{red}{\textbf{80.37 ($\pm$10.74)}} & \textcolor{blue}{\textbf{93.08 ($\pm$7.20)}} &  \textcolor{red}{\textbf{0.90 ($\pm$0.36)}}\\
\midrule
\end{tabular}

\caption{Comparison to other state-of-the-art architectures on the \textcolor{blue}{four analyzed} datasets. Best and second best results are represented in red and blue bold, respectively. }
\label{table:sota_comp}
\end{table}

\paragraph*{\textbf{Qualitative evaluation}}

\begin{figure*}[h!]
    \centering
    \includegraphics[width=1\textwidth]{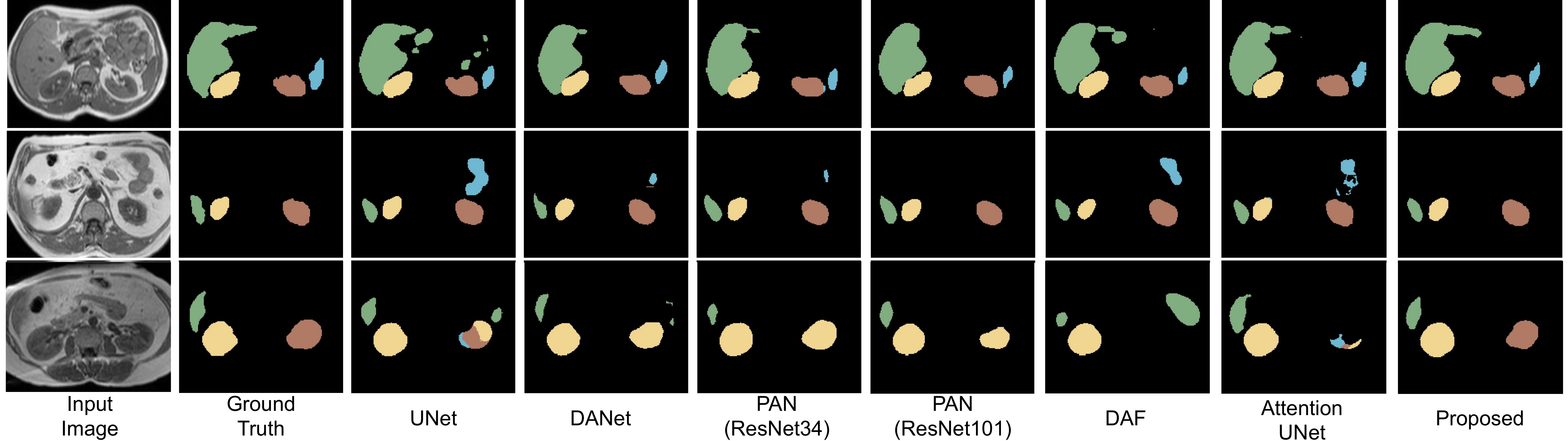}
    \caption{Results on \textcolor{black}{three} subjects on the CHAOS Challenge dataset. The proposed multi-scale guided attention network achieves qualitatively better results than other state-of-the-art networks that also integrate attention modules.}
        \label{fig:visualImages}
\end{figure*}

To visualize the impact of the different attention modules,  Fig. \ref{fig:visualImages} displays \textcolor{black}{the segmentation results on three CHAOS subjects.} Despite the similar results reported on Table \ref{table:sota_comp} for several architectures, the qualitative results depict interesting findings. First, we can observe that UNet typically under-segments certain organs and gets confused easily. For example, in the second row it confused the small bowels with the spleen, while the spleen is not even present in that slice. Integrating attention can overcome some of these limitations and improve the segmentation performance by focusing the attention to relevant areas. This can be observed in the results obtained by the other networks, which, up to some extent, reduce the amount of false positives. Nevertheless, it produces smoother segmentations, resulting in a loss of fine grained details. An interesting result is the segmentation in the last row, where all the models except the proposed network get confused to segment the left kidney. While DANet and PAN models confuse the left kidney with the right one, DAF is not able to detect any relevant region related to the kidneys in that area. In addition, both UNet and UNet with attention models generate segmentations of the left kidney that contain three organs, i.e., left and right kidneys and spleen, which is anatomically not plausible. Unlike all these models, the proposed architecture does not get distracted by ambiguous regions and some misclassified structures are now correctly classified. 

\begin{figure*}[h!]
    \centering
    \includegraphics[width=1\textwidth]{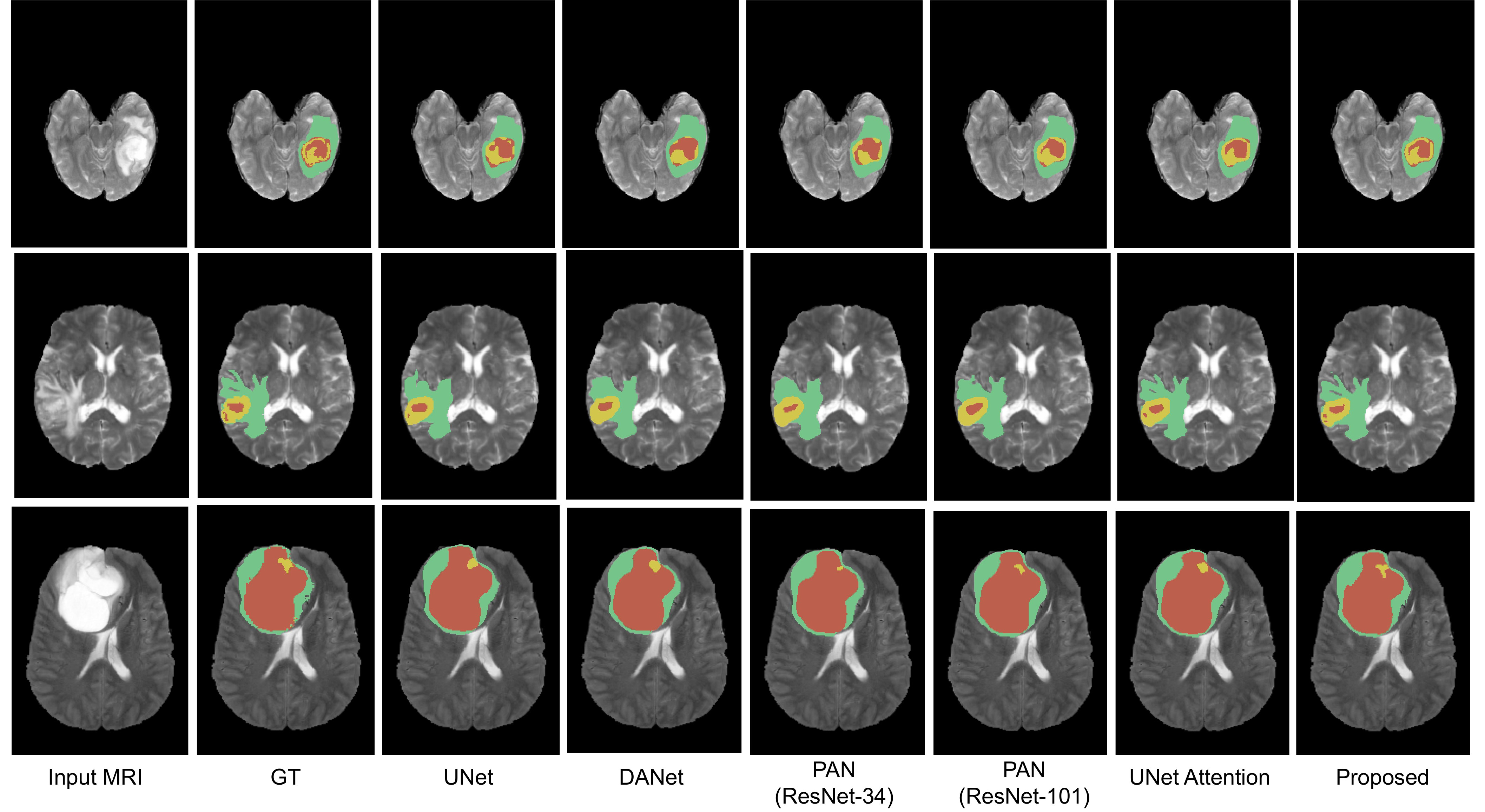}
    \caption{\textcolor{black}{Results on three subjects on the BRATS Challenge dataset. In these figures, the following tumor structures are depicted: oedema (green), enhancing core (yellow) and necrotic or tumor core (red).}}
        \label{fig:visualImagesBRATS}
\end{figure*}

\textcolor{black}{Similar results are observed on the segmentations obtained in the BRATS'18 images (Fig. \ref{fig:visualImagesBRATS}). Particularly, we can see that the proposed network obtains finer details than the other architectures. For example, small ramifications on the oedema (in green) are better captured by the proposed model (\textit{second row}). Likewise, segmentation of necrotic areas (in red) achieved by our method is closer to the ground truth, specially when the region has a complex shape (\textit{first row}).} These visual results indicate that our approach can successfully recover finer segmentation details, while avoiding getting distracted in ambiguous regions. The selective integration of spatial information and among channel maps followed by a guided attention module helps to capture context information. This demonstrates that the proposed multi-scale guided attention model can efficiently encode complimentary information to accurately segment medical images. 

\paragraph*{\textbf{Visual inspection of feature maps}}

\begin{figure}[h!]
    \centering
    \includegraphics[width=0.5\textwidth]{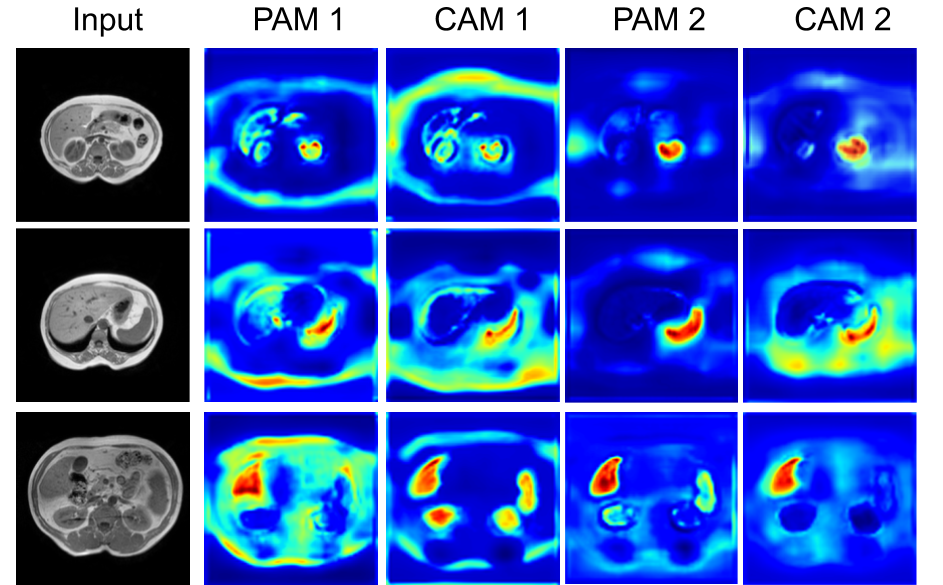}
    \caption{Visualization results of the channel maps. For each row, we show an input image, and the corresponding channel maps from the outputs of spatial (PAM) and channel (CAM) attention module at guided module of the Fig. \ref{fig:guidedModule}.}
    \label{fig:attmaps}
\end{figure}

Showing the performance difference through ablation studies and quantitative evaluations alone may not be enough to fully understand the benefits and behaviour of novel models. Although the proposed modules contribute to the performance improvement, as shown in the results, it is interesting to investigate whether different modules work as expected. To this end, we analyze some attended feature maps from both the spatial and channel attention modules \textcolor{black}{(Fig. \ref{fig:attmaps})}. We find that the response of specific semantic classes is more noticeable after the second guided attention modules, i.e., PAM 2 and CAM 2 attentive features. While spatial and channel attention can highlight specific class semantics in the first step of the guided module (second and third column), some non-targeted regions are still highlighted on the semantic maps. Furthermore, highest values are also more spread over the entire image. Contrary, the proposed guided attention module generates features (fourth and fifth columns) that better focus on the specific regions of the structures of interest. Particularly, it can be observed that there exist feature maps whose highlighted areas concentrate on a single organ, avoiding ambiguous regions that might result on misclassification of some regions.

\section{Conclusion}

In this work, we introduced a novel attention architecture for the task of medical image segmentation. This model incorporates a multi-scale strategy to combine semantic information at different levels and self-attention modules to progressively aggregate relevant contextual features. Last, a guided refinement module filters noisy regions and help the network to focus on relevant class-specific regions in the image. To validate our approach we conducted experiments on \textcolor{black}{ three different segmentation tasks: abdominal organ, cardiovascular structures and brain tumor.} We provided extensive experiments to evaluate the impact of the individual components of the proposed architecture. Besides, we compared our model to existing approaches that integrate attention, which have been recently proposed for natural scene \cite{fu2018dual,li2018pyramid} and medical image \cite{ronneberger2015u,wang18d,schlemper2019attention} segmentation. Experiment results showed that the proposed model outperformed all previous approaches both quantitative and qualitatively, which may be explained by the enhanced ability to model rich contextual dependencies over local features. This demonstrates the efficiency of our approach to provide precise and reliable automatic segmentations of medical images.

\section*{Acknowledgments}

We wish to thank NVIDIA for its kind donation of the Titan V GPU used in this work.


\bibliographystyle{IEEEtran}
\bibliography{refs_shorted}

\clearpage

\setcounter{page}{1}
\begin{center}
\textbf{\large Supplemental Materials}
\end{center}

\subsection*{\textbf{Evaluation metrics: formulation}}

In this section, we give the formal definition of the metrics employed to evaluate the proposed architecture.
\paragraph{\textbf{Dice Similarity Coefficient (DSC)}}
Given two volumes $A$ and $B$, their DSC can be defined as:

\begin{equation}
	\mathrm{DSC} \ = \ \dfrac{2\left|A \cap B \right|}{|A|+|B|}
\end{equation}

In this metric, values close to 1 indicate high degree of overlapping, whereas near 0 represent not overlapping at all. 
\paragraph{\textbf{Volume Similarity (VS)}} Further, we also assess the segmentation performance based on the volume similarity, which is formulated as:

\begin{equation}
	\mathrm{VS} \ = 1 - abs(A - B) / (A + B)
\end{equation}

\paragraph{\textbf{Mean Surface Distance (MSD)}} The MSD between contours $A$ and $B$ is defined as follows:
\begin{equation}
\mathrm{MSD} \ = \ \dfrac{1}{|A| + |B|} \, \left(\sum_{a \in A} \min_{b \in B} \, d(a,b) + \sum_{b \in B} \min_{a \in A} \, d(b,a) \right)
\end{equation}

\begin{equation}
\mathrm{MSD} \ = \ \dfrac{1}{|A| + |B|} \, \left(\sum_{a \in A} d(a,b) + \sum_{b \in B}  \, d(b,a) \right)
\end{equation}

where $d(a,b)$ is the distance between a point $a$ on the surface A and the surface B, which is given by the minimum of the Euclidean norm:

\begin{equation}
    d(a,B) = \min_{b \in B} \| a - b\|^2_2
\end{equation}


\subsection*{\textbf{Additional results}}

\begin{table*}[ht!]
\centering
\scriptsize
\begin{tabular}{lcccc|c}\\
\toprule
 & \multicolumn{5}{c}{\textbf{DSC} (\%)}\\
 \midrule
Method & \textbf{Liver} & \textbf{Kidney R} & \textbf{Kidney L} & \textbf{Spleen} & \textbf{Mean}  \\
\midrule
Baseline (DAF \cite{wang18d})  &  91.66 ($\pm$2.99) & 79.28 ($\pm$18.68)& 83.63 ($\pm$7.56) &  75.35 ($\pm$20.41)&  82.48 ($\pm$6.06)   \\
Proposed (PAM)     &  91.89 ($\pm$4.29)& 85.47 ($\pm$7.04)& 86.84 ($\pm$6.53)  & 73.65 ($\pm$22.62)& 84.46 ($\pm$6.68)   \\
Proposed (CAM)     &  92.58 ($\pm$2.65) & 84.52 ($\pm$9.34)& 86.38 ($\pm$6.27)& 76.84 ($\pm$20.56) & 85.08 ($\pm$5.62)\\
Proposed (DANet) &  \textcolor{blue}{\textbf{92.60 ($\pm$3.20)}} & 85.29 ($\pm$7.96) & 87.74 ($\pm$6.37)& 76.44 ($\pm$22.17)&  85.52 ($\pm$5.86)       \\
Proposed (MS-Dual)  & \textcolor{red}{\textbf{92.62 ($\pm$3.08)}}  & \textcolor{blue}{\textbf{86.29 ($\pm$5.98)}} & \textcolor{red}{\textbf{88.82 ($\pm$4.84)}}  & \textcolor{blue}{\textbf{76.96 ($\pm$19.87)}} &  \textcolor{blue}{\textbf{86.17 ($\pm$5.78)}}\\
Proposed (MS-Dual-Guided)  & 92.46 ($\pm$2.82)  & \textcolor{red}{\textbf{87.96 ($\pm$6.46)}}& \textcolor{blue}{\textbf{88.01 ($\pm$6.16)}}  & \textcolor{red}{\textbf{78.61 ($\pm$18.69)}} &  \textcolor{red}{\textbf{86.75 ($\pm$5.05)}}\\
\midrule
& \multicolumn{5}{c}{\textbf{Volume similarity (VS)} (\%)}\\
 \midrule
 & \textbf{Liver} & \textbf{Kidney R} & \textbf{Kidney L} & \textbf{Spleen} & \textbf{Mean}  \\

\midrule
Proposed( DAF \cite{wang18d})  &  96.69 ($\pm$3.21) & 86.75 ($\pm$16.41) & 90.29 ($\pm$8.39) & \textcolor{blue}{\textbf{84.98 ($\pm$14.42)}} &   89.68 ($\pm$4.48)  \\
Proposed (PAM)     & 96.62 ($\pm$4.62)& 92.83 ($\pm$7.43)& 93.96 ($\pm$6.46)& 83.93 ($\pm$20.54)&  91.84 ($\pm$4.77) \\
Proposed (CAM)     & \textcolor{blue}{\textbf{97.25 ($\pm$2.95)}}  & 93.78 ($\pm$6.04)& 93.98 ($\pm$5.48)& 83.72 ($\pm$20.97)& 92.18 ($\pm$5.07)\\
Proposed (DANet) &  97.04 ($\pm$3.03)& \textcolor{blue}{\textbf{94.50 ($\pm$5.96)}} & 93.43 ($\pm$7.03)& 83.30 ($\pm$22.53)&  92.07 ($\pm$5.23)     \\
Proposed (MS-Dual)  & \textcolor{red}{\textbf{97.47 ($\pm$3.07)}}  & 93.30 ($\pm$4.11)& \textcolor{red}{\textbf{95.27 ($\pm$4.89)}}  & 84.90 ($\pm$16.86) & \textcolor{blue}{\textbf{92.74 ($\pm$4.76)}} \\
Proposed (MS-Dual-Guided)  &  96.44 ($\pm$3.15)  & \textcolor{red}{\textbf{96.14 ($\pm$3.15)}}& \textcolor{blue}{\textbf{94.95 ($\pm$4.48)}}  & \textcolor{red}{\textbf{87.87 ($\pm$15.23)}} & \textcolor{red}{\textbf{93.85 ($\pm$3.50)}} \\

\midrule
& \multicolumn{5}{c}{\textbf{Average Surface Distance (MSD)} (voxels)}\\
 \midrule
 & \textbf{Liver} & \textbf{Kidney R} & \textbf{Kidney L} & \textbf{Spleen} & \textbf{Mean}  \\
 \midrule
Baseline( DAF \cite{wang18d})  & 0.64 ($\pm$0.29) & 0.97 ($\pm$1.08) & 0.63 ($\pm$0.25)& 1.45 ($\pm$2.04)&  0.92 ($\pm$0.33)  \\
Proposed (PAM)     & 0.55 ($\pm$0.19)& 0.56 ($\pm$0.23)& 0.55 ($\pm$0.21)& 1.54 ($\pm$2.40)&  0.80 ($\pm$0.43) \\
Proposed (CAM)     &  0.58 ($\pm$0.22)& 0.57 ($\pm$0.24)& 0.52 ($\pm$0.20)& 1.29 ($\pm$1.64)  & 0.74 ($\pm$0.32)\\
Proposed (DANet) & \textcolor{blue}{\textbf{0.54 ($\pm$0.19)}}  &0.56 ($\pm$0.19) & 0.50 ($\pm$0.18)& 1.49 ($\pm$2.29)&   0.77 ($\pm$0.41) \\
Proposed (MS-Dual)  & \textcolor{red}{\textbf{0.53 ($\pm$0.18)}}  & \textcolor{blue}{\textbf{0.51 ($\pm$0.14)}}&  \textcolor{red}{\textbf{0.46 ($\pm$0.14)}} & \textcolor{blue}{\textbf{1.19 ($\pm$1.42)}} &  \textcolor{blue}{\textbf{0.67 ($\pm$0.30)}}\\
Proposed (MS-Dual-Guided)  & \textcolor{blue}{\textbf{0.54 ($\pm$0.16)}}  & \textcolor{red}{\textbf{0.48 ($\pm$0.18)}}&  \textcolor{blue}{\textbf{0.48 ($\pm$0.14)}} & \textcolor{red}{\textbf{1.13 ($\pm$1.24)}} &  \textcolor{red}{\textbf{0.66 ($\pm$0.27)}}\\
\midrule


\midrule
\end{tabular}
\caption{Ablation study on different proposed attention modules on the Chaos dataset (multi-organ segmentation on MRI task). The values show the average result of the experiments averaged over the 3 folds. Best results are represented in red bold, while blue is used to highlight the second best performance.}
\label{table:metrics_ablation_supplemental}
\end{table*}

\begin{table*}[t!]
\centering
\scriptsize
\begin{tabular}{lcccc|c}\\
\toprule
 & \multicolumn{5}{c}{\textbf{DSC} (\%)}\\
 \midrule
Method & \textbf{Liver} & \textbf{Kidney R} & \textbf{Kidney L} & \textbf{Spleen} & \textbf{Mean}  \\
 \midrule
UNet \cite{ronneberger2015u}  & 90.94 ($\pm$4.01)  & 79.14 ($\pm$15.23) & 82.51 ($\pm$7.48) & 71.95 ($\pm$21.61) &  81.14 ($\pm$7.88) \\
DANet \cite{fu2018dual}  & 91.69  ($\pm$4.07) & 83.85 ($\pm$9.40)& 84.49 ($\pm$8.60) & 75.54 ($\pm$16.08)&  83.89 ($\pm$9.54)   \\
PAN (ResNet34) \cite{li2018pyramid}   & 91.99 ($\pm$2.98)  & 81.51 ($\pm$9.03) & 83.62 ($\pm$6.21) & 73.70 ($\pm$19.97) & 82.70 ($\pm$6.51) \\
PAN (ResNet101)\cite{li2018pyramid}   & \textcolor{blue}{\textbf{92.13 ($\pm$3.51)}}  & \textcolor{blue}{\textbf{85.02 ($\pm$5.16)}} & 85.36 ($\pm$4.87) & 74.84 ($\pm$21.23) & 84.34 ($\pm$6.17) \\
DAF \cite{wang18d}  &  91.66 ($\pm$2.99) & 79.28 ($\pm$18.68)& 83.63 ($\pm$7.56) &  75.35 ($\pm$20.41)&  82.48 ($\pm$6.06)   \\
UNet Attention \cite{schlemper2019attention}  &   92.02 ($\pm$1.93) & 84.33 ($\pm$5.91)&  \textcolor{blue}{\textbf{85.57 ($\pm$4.09)}} & \textcolor{blue}{\textbf{77.18 ($\pm$15.95)}} &  \textcolor{blue}{\textbf{84.77 ($\pm$5.27)}}   \\
Proposed (MS-Dual-Guided)  & \textcolor{red}{\textbf{92.46 ($\pm$2.82)}}  & \textcolor{red}{\textbf{87.96 ($\pm$6.46)}}& \textcolor{red}{\textbf{88.01 ($\pm$6.16)}}  & \textcolor{red}{\textbf{78.61 ($\pm$18.69)}} &  \textcolor{red}{\textbf{86.75 ($\pm$5.05)}}\\

\midrule
& \multicolumn{5}{c}{\textbf{Volume similarity (VS)} (\%)}\\
 \midrule
 & \textbf{Liver} & \textbf{Kidney R} & \textbf{Kidney L} & \textbf{Spleen} & \textbf{Mean}  \\
 \midrule
UNet \cite{ronneberger2015u}  &  95.54 ($\pm$4.43)& 87.68 ($\pm$5.77)& 89.55 ($\pm$4.68)& 83.28 ($\pm$14.78)& 89.01 ($\pm$4.82)\\
DANet \cite{fu2018dual}  &  96.90 ($\pm$4.18) & 92.88 ($\pm$5.12)& 91.52 ($\pm$6.73) & 84.37 ($\pm$16.15)& 91.42  ($\pm$4.52)   \\
PAN (ResNet34) \cite{li2018pyramid}   &  96.56 ($\pm$3.55)  & 90.89 ($\pm$5.64) & 91.83 ($\pm$7.75) & 81.98 ($\pm$20.67) & 90.32 ($\pm$5.27) \\
PAN (ResNet101) \cite{li2018pyramid}   & \textcolor{red}{\textbf{96.99 ($\pm$3.64)}}  & \textcolor{blue}{\textbf{93.77 ($\pm$4.63)}} & \textcolor{blue}{\textbf{92.69 ($\pm$6.88)}} & 84.24 ($\pm$17.37) & \textcolor{blue}{\textbf{91.93 ($\pm$4.71)}} \\
DAF \cite{wang18d}  &  96.69 ($\pm$3.21) & 86.75 ($\pm$16.41) & 90.29 ($\pm$8.39) & 84.98 ($\pm$14.42) &   89.68 ($\pm$4.48)  \\
UNet Attention \cite{schlemper2019attention}  &   \textcolor{blue}{\textbf{96.95 ($\pm$1.89)}} & 92.29 ($\pm$6.41)& 91.79 ($\pm$3.53) & \textcolor{blue}{\textbf{85.94 ($\pm$11.88)}} &  91.74 ($\pm$3.91)   \\
Proposed (MS-Dual-Guided)  &  96.44 ($\pm$3.15)  & \textcolor{red}{\textbf{96.14 ($\pm$3.15)}}& \textcolor{red}{\textbf{94.95 ($\pm$4.48)}}  & \textcolor{red}{\textbf{87.87 ($\pm$15.23)}} & \textcolor{red}{\textbf{93.85 ($\pm$3.50)}} \\
\midrule
& \multicolumn{5}{c}{\textbf{Average Surface Distance (MSD)} (voxels)}\\
 \midrule
 & \textbf{Liver} & \textbf{Kidney R} & \textbf{Kidney L} & \textbf{Spleen} & \textbf{Mean}  \\
 \midrule
UNet \cite{ronneberger2015u}  & 0.59 ($\pm$0.18)& 0.69 ($\pm$0.38)& 0.61 ($\pm$0.19)& 1.76 ($\pm$2.57)& 0.91 ($\pm$0.49) \\
DANet \cite{fu2018dual}  &  0.61 ($\pm$0.27) & 0.65 ($\pm$0.31)& 0.67 ($\pm$0.30) & 1.17 ($\pm$0.94)& 0.78  ($\pm$0.23)   \\
PAN (ResNet34)\cite{li2018pyramid}   & 0.62 ($\pm$0.25)  & 0.75 ($\pm$0.31) & 0.69 ($\pm$0.21) & 1.37 ($\pm$1.43) & 0.86 ($\pm$0.29) \\
PAN (ResNet101) \cite{li2018pyramid}   & \textcolor{blue}{\textbf{0.57 ($\pm$0.22)}}  & \textcolor{blue}{\textbf{0.61 ($\pm$0.19)}} & 0.64 ($\pm$0.15) & 1.30 ($\pm$1.47) & 0.78 ($\pm$0.31) \\
DAF \cite{wang18d}  & 0.64 ($\pm$0.29) & 0.97 ($\pm$1.08) & 0.63 ($\pm$0.25)& 1.45 ($\pm$2.04)&  0.92 ($\pm$0.33)  \\
UNet Attention \cite{schlemper2019attention}  &   \textcolor{blue}{\textbf{0.57 ($\pm$0.25)}} & \textcolor{blue}{\textbf{0.61 ($\pm$0.23)}}&  \textcolor{blue}{\textbf{0.56 ($\pm$0.18)}} & \textcolor{blue}{\textbf{1.15 ($\pm$1.01)}}&  \textcolor{blue}{\textbf{0.72 ($\pm$0.24)}}   \\
Proposed (MS-Dual-Guided)  & \textcolor{red}{\textbf{0.54 ($\pm$0.16)}}  & \textcolor{red}{\textbf{0.48 ($\pm$0.18)}}&  \textcolor{red}{\textbf{0.48 ($\pm$0.14)}} & \textcolor{red}{\textbf{1.13 ($\pm$1.24)}} &  \textcolor{red}{\textbf{0.66 ($\pm$0.27)}} \\
\midrule
\midrule
\end{tabular}

\caption{Comparison of the proposed network to other state-of-the-art architectures on the CHAOS dataset (multi-organ segmentation on MRI task). The values show the average result of the experiments averaged over the 3 folds. Best results are represented in red bold, while blue is used to highlight the second best performance.}
\label{table:sota_comp_supplemental}
\end{table*}

Tables \ref{table:metrics_ablation_supplemental}, \ref{table:sota_comp_supplemental} and \ref{table:sota_comp_supplemental_hsvm} report the extended version of the experimental results on the ablation study and comparison to other state-of-the-art networks. In these tables, individual results on single organs are also included to provide the reader a wider view of the performance of the different methods. We can observe that the proposed architecture is consistently outperforming other models, ranking either first or second in almost all the organs for all the evaluation metrics. The only exception is the result obtained for liver segmentation in terms of volume similarity, where all the models obtain almost identical results.

\newcolumntype{L}{>{\arraybackslash}p{2.3cm}}

\begin{table}[t!]
\centering
\scriptsize
\begin{tabular}{Lcc|c}\\
\toprule
 & \multicolumn{3}{c}{\textbf{DSC}}\\
 \midrule
Method & \textbf{Myocardium} & \textbf{Blood Pool} &\textbf{Mean}  \\
 \midrule
UNet \cite{ronneberger2015u}  & 71.77 ($\pm$9.36)  & 87.84 ($\pm$4.35) & 79.80 ($\pm$6.72)  \\

DANet \cite{fu2018dual}  & \textcolor{blue}{\textbf{75.85 ($\pm$9.10)}} & 89.24 ($\pm$3.56)&  \textcolor{blue}{\textbf{82.55 ($\pm$5.91)}} \\

PAN (ResNet34) \cite{li2018pyramid}   & 72.90 ($\pm$11.93) & 89.04 ($\pm$3.69) &  80.97 ($\pm$7.76) \\
PAN (ResNet101)\cite{li2018pyramid}   & 74.98 ($\pm$7.68)  & \textcolor{red}{\textbf{89.53 ($\pm$2.97)}} & 82.26 ($\pm$5.08) \\
DAF \cite{wang18d}  &  74.08 ($\pm$8.55) & \textcolor{blue}{\textbf{89.48 ($\pm$3.39)}}& 81.78 ($\pm$5.71)    \\

UNet Attention \cite{schlemper2019attention}  &   74.50 ($\pm$10.13) & 88.66 ($\pm$4.25)&  81.58 ($\pm$6.84)  \\

Proposed  & \textcolor{red}{\textbf{77.10 ($\pm$6.94)}}  & 89.30 ($\pm$3.50)& \textcolor{red}{\textbf{83.20 ($\pm$4.93)}}\\

\midrule
& \multicolumn{3}{c}{\textbf{Volume similarity (VS)}}\\
 \midrule
 & \textbf{Myocardium} & \textbf{Blood Pool} &  \textbf{Mean}  \\
 \midrule
UNet \cite{ronneberger2015u}  &  91.05 ($\pm$9.75)& 95.78 ($\pm$4.04)& 93.41 ($\pm$6.44)\\

DANet \cite{fu2018dual}  &  91.80 ($\pm$8.95) & \textcolor{red}{\textbf{97.50 ($\pm$3.01)}}& \textcolor{red}{\textbf{94.65 ($\pm$4.45)}}  \\

PAN (ResNet34) \cite{li2018pyramid}   & 90.58 ($\pm$10.89) &  96.93 ($\pm$3.66) & 93.76 ($\pm$5.85) \\

PAN (ResNet101) \cite{li2018pyramid}   & 91.42 ($\pm$7.59) & \textcolor{blue}{\textbf{97.23 ($\pm$2.36)}} & 94.33 ($\pm$3.69)\\
DAF \cite{wang18d}  &  91.73 ($\pm$6.30) & 96.89 ($\pm$2.33) & 94.31 ($\pm$3.21)  \\

UNet Attention \cite{schlemper2019attention}  &   \textcolor{red}{\textbf{92.52 ($\pm$7.66)}} & 96.69 ($\pm$2.20)& \textcolor{blue}{\textbf{94.61 ($\pm$4.17)}}   \\

Proposed  &  \textcolor{blue}{\textbf{92.08 ($\pm$4.39)}}  & 96.82 ($\pm$2.76)& 
94.45 ($\pm$2.39)  \\

\midrule
& \multicolumn{3}{c}{\textbf{Average Surface Distance (MSD)}}\\
 \midrule
 & \textbf{Myocardium} & \textbf{Blood pool} &  \textbf{Mean}  \\
 \midrule
UNet \cite{ronneberger2015u}  & 1.82 ($\pm$1.48)& 1.55 ($\pm$1.08)& 1.68 ($\pm$1.28)\\

DANet \cite{fu2018dual}  &  \textcolor{blue}{\textbf{1.23 ($\pm$0.51)}} & 1.32 ($\pm$0.46)& 1.27 ($\pm$0.46)   \\

PAN (ResNet34)\cite{li2018pyramid}   & 1.97 ($\pm$1.84) & 1.26 ($\pm$0.48) & 1.62 ($\pm$1.19)\\

PAN (ResNet101) \cite{li2018pyramid}   & 1.33 ($\pm$0.53)  & \textcolor{red}{\textbf{1.15 ($\pm$0.30)}} & 1.24 ($\pm$0.38)   \\
DAF \cite{wang18d}  & 1.41 ($\pm$0.45) & 1.44 ($\pm$0.46) & 1.48 ($\pm$0.50)  \\

UNet Attention \cite{schlemper2019attention} & 1.24 ($\pm$0.42) & 1.25 ($\pm$0.39)  & \textcolor{blue}{\textbf{1.25 ($\pm$0.42)}} \\

Proposed  & \textcolor{red}{\textbf{1.15 ($\pm$0.33)}}  & \textcolor{blue}{\textbf{1.24 ($\pm$0.43)}}&  \textcolor{red}{\textbf{1.19 ($\pm$0.37)}} \\

\midrule
\midrule
\end{tabular}

\caption{Comparison of the proposed network to other state-of-the-art architectures on the HVSMR 2016 dataset. The values show the average result of the experiments on the 5 folds. }
\label{table:sota_comp_supplemental_hsvm}
\end{table}

\begin{figure}[h!]
\subfloat[Dice Similarity coefficient (\%)]{\includegraphics[width =1.1 \linewidth]{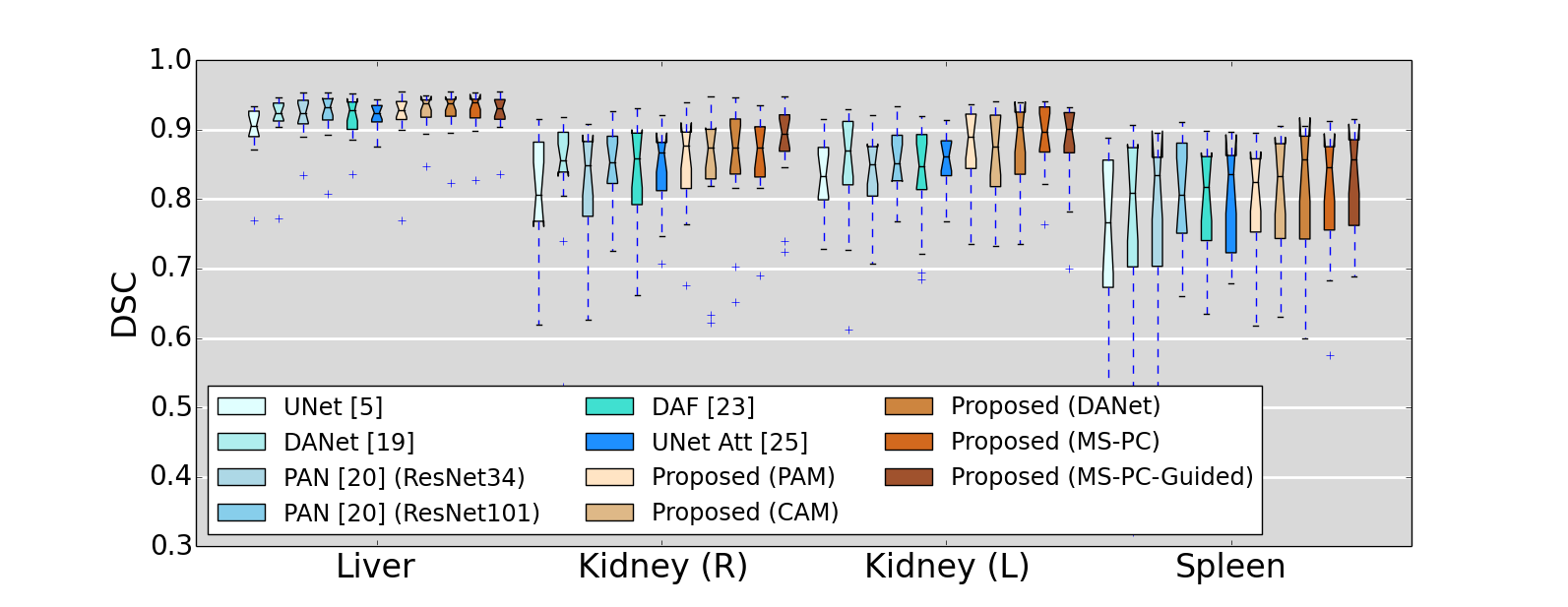}} \\
\subfloat[Volume similarity (\%)]{\includegraphics[width = 1.1 \linewidth]{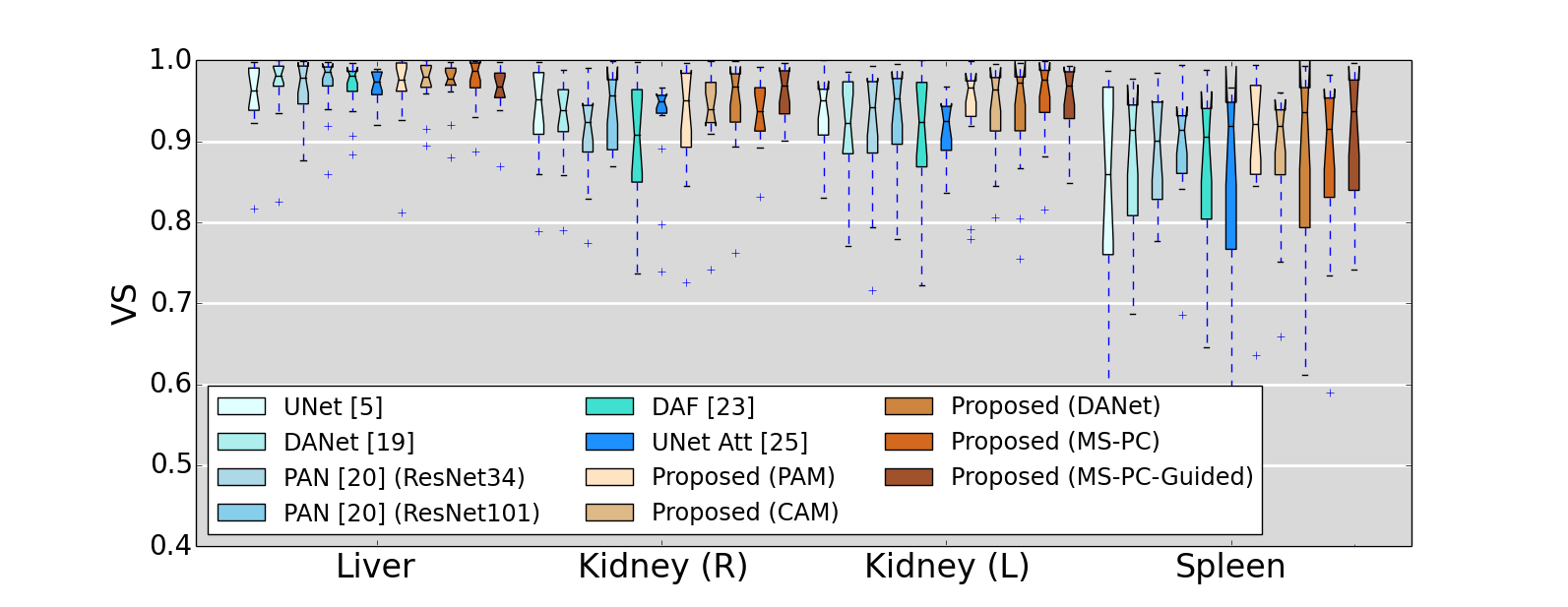}}\\
\subfloat[Average surface distance (voxels)]{\includegraphics[width = 1.1 \linewidth]{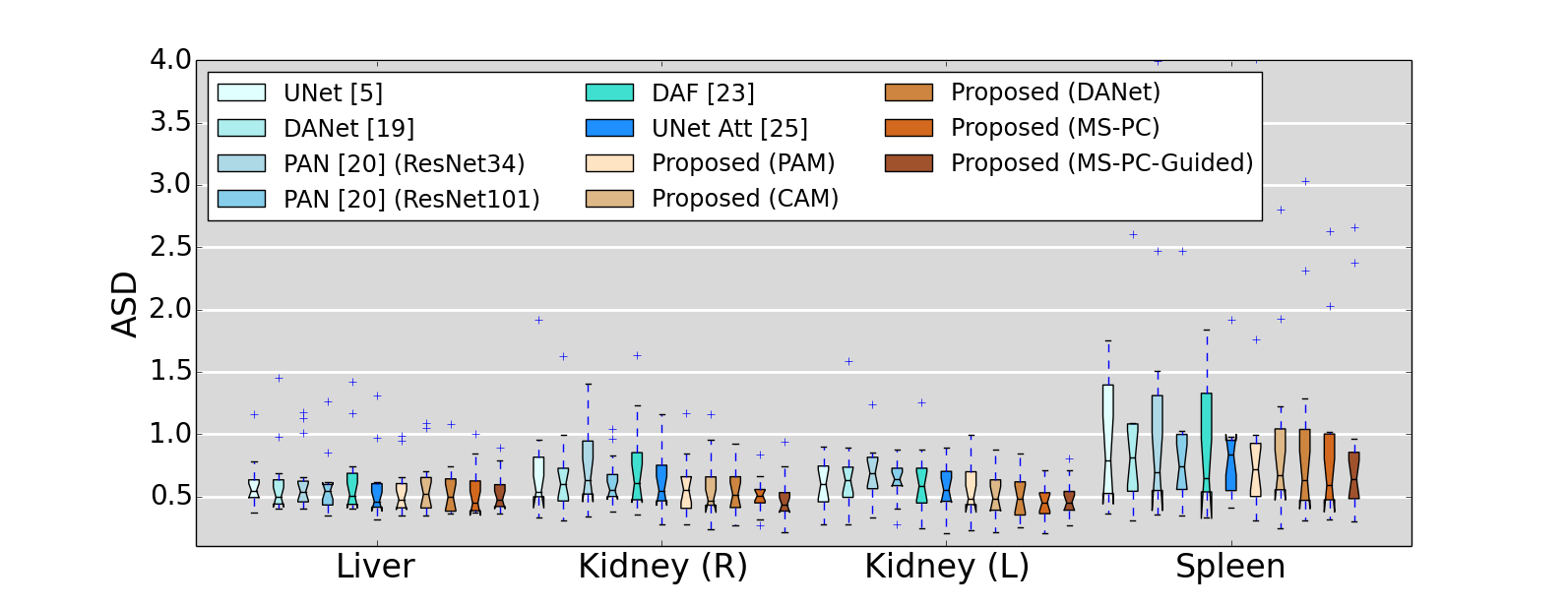}} 
    \caption{These plots depict the distributions of the different evaluation metrics for the four organs segmented. Bluish colors represent the results obtained by other state-of-the-art networks, whereas the results obtained by our proposed models are displayed in with the brownish boxplots.}
    \label{fig:metrics}
\end{figure}

In addition to the values reported on Tables \ref{table:metrics_ablation_supplemental} and \ref{table:sota_comp_supplemental} in the Supplemental Material, we also depict the distribution of DSC, VS and MSD values on the 15 subjects used for evaluation for all the models (Fig. \ref{fig:metrics}). In these plots, we can first observe the impact of the different attention modules in the segmentation performance of the proposed model. As we progressively include the proposed attention modules in the baseline network, the segmentation performance improves, which is reflected in a better distribution of segmentation accuracy values with a smaller variance. This difference on results distribution is more prominent when comparing the proposed network with other state-of-the-art networks, which are represented in bluish box plots. We can also observe that this pattern is constant across organs and metrics, suggesting that the proposed attention network achieves better and more robust segmentation results than current state-of-the-art architectures.

\subsubsection{\textbf{Convergence}}

\begin{figure}[h!]
    \centering
    \includegraphics[width=.525\textwidth]{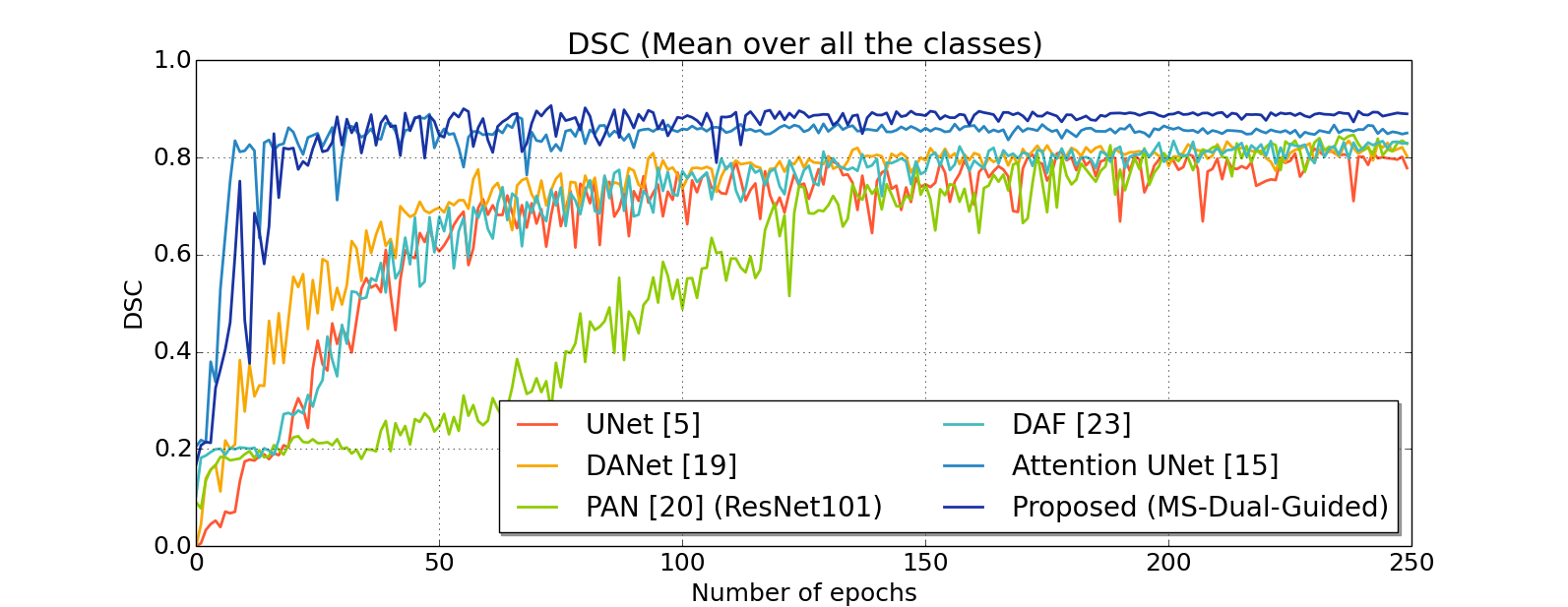}
    \caption{Evolution of the mean validation DSC over time.}
    \label{fig:DSC_Evolution}
\end{figure}

We have also compared the different architectures in terms of convergence, whose results are depicted in Fig. \ref{fig:DSC_Evolution}. Particularly, the mean DSC value over the four structures on one of the validation folds is shown for each of the networks. It can be observed that, even though most of the networks achieve results which may be considered `similar' --up to some extent-- the convergence behaviour is totally different. While there are three networks with similar convergence curves --i.e., UNet, DANet and DAF--, PAN needs more iterations to convergence, ultimately performing better than these networks after nearly 400 epochs. On the other hand, we found that attention UNet and the proposed network presented the fastest convergence, achieving their best results at epoch 48 and 73, respectively.

\begin{table*}[t!]
\centering
\scriptsize
\begin{tabular}{lcccc|c}\\
\toprule
 & \multicolumn{5}{c}{\textbf{DSC} (\%)}\\
 \midrule
Method & \textbf{ED} & \textbf{ET} & \textbf{TC} & \textbf{--} & \textbf{Mean}  \\
 \midrule
UNet \cite{ronneberger2015u}  & 84.87 ($\pm$6.82)  & 56.38 ($\pm$27.55) & 79.71 ($\pm$11.70) & -- &  73.65 ($\pm$12.39) \\
DANet \cite{fu2018dual}  & \textcolor{blue}{\textbf{88.24  ($\pm$5.39)}} & 63.69 ($\pm$22.25)& \textcolor{blue}{\textbf{85.33 ($\pm$6.92)}} & --&  \textcolor{blue}{\textbf{79.09 ($\pm$10.89)}}   \\
PAN (ResNet34) \cite{li2018pyramid}   & 85.25 ($\pm$6.64)  & 55.89 ($\pm$27.76) & 81.23  ($\pm$8.22) & -- & 74.12 ($\pm$12.76) \\
PAN (ResNet101)\cite{li2018pyramid}   & 87.07 ($\pm$6.67)  & 60.77 ($\pm$24.74) & 82.82 ($\pm$8.76) & -- & 76.89 ($\pm$11.53) \\
DAF \cite{wang18d}  &  86.87 ($\pm$5.94) & 60.28 ($\pm$24.74)& 83.18 ($\pm$8.39) &  -- &  76.78 ($\pm$11.77)   \\
UNet Attention \cite{schlemper2019attention}  &   87.50 ($\pm$5.66) & \textcolor{blue}{\textbf{63.74 ($\pm$22.65)}}&  84.59 ($\pm$7.43) & -- &  78.61 ($\pm$10.58)   \\
Proposed (MS-Dual-Guided)  & \textcolor{red}{\textbf{89.11 ($\pm$4.94)}}  & \textcolor{red}{\textbf{65.25 ($\pm$.2285)}}& \textcolor{red}{\textbf{86.76 ($\pm$6.49)}}  & -- &  \textcolor{red}{\textbf{80.37 ($\pm$10.74)}} \\

\midrule
& \multicolumn{5}{c}{\textbf{Volume similarity (VS)} (\%)}\\
 \midrule
 & \textbf{ED} & \textbf{ET} & \textbf{TC} & \textbf{--} & \textbf{Mean}  \\
 \midrule
UNet \cite{ronneberger2015u}  &  96.36 ($\pm$4.08)& 75.81 ($\pm$27.23)& 90.99  ($\pm$11.63)& --& 87.72 ($\pm$8.70)\\
DANet \cite{fu2018dual}  &  \textcolor{red}{\textbf{99.04 ($\pm$1.21)}} & \textcolor{red}{\textbf{83.47 ($\pm$20.11)}}& \textcolor{blue}{\textbf{97.45 ($\pm$2.95)}} & --& \textcolor{red}{\textbf{93.32 ($\pm$6.99)}}  \\
PAN (ResNet34) \cite{li2018pyramid}   &  98.05 ($\pm$1.98)& 75.87 ($\pm$28.17)& 95.63 ($\pm$4.41)& -- & 89.85 ($\pm$9.93)  \\
PAN (ResNet101) \cite{li2018pyramid}   &  \textcolor{blue}{\textbf{98.68 ($\pm$2.21)}} & 80.38 ($\pm$24.83)& 96.22 ($\pm$5.89) & -- & 91.76 ($\pm$8.11)\\
DAF \cite{wang18d}  &  97.99 ($\pm$2.10) & 77.86 ($\pm$24.92) & 95.88 ($\pm$5.26) & -- &   90.58 ($\pm$9.03)  \\
UNet Attention \cite{schlemper2019attention}  &   98.14 ($\pm$1.88) & \textcolor{blue}{\textbf{82.99 ($\pm$21.09)}}& 96.84 ($\pm$2.87) & -- &  92.66 ($\pm$6.86)  \\
Proposed (MS-Dual-Guided)  & 98.54 ($\pm$1.76) & 82.91 ($\pm$20.17) &  \textcolor{red}{\textbf{97.78 ($\pm$2.56)}} &-- & \textcolor{blue}{\textbf{93.08 ($\pm$7.20)}} \\
\midrule
& \multicolumn{5}{c}{\textbf{Average Surface Distance (MSD)} (voxels)}\\
 \midrule
 & \textbf{ED} & \textbf{ET} & \textbf{TC} & \textbf{-} & \textbf{Mean}  \\
 \midrule
UNet \cite{ronneberger2015u}  & 0.99 ($\pm$0.33)& 2.37 ($\pm$1.74)& 1.56 ($\pm$1.34)& -- & 1.65 ($\pm$0.57) \\
DANet \cite{fu2018dual}  &  \textcolor{blue}{\textbf{0.67 ($\pm$0.16)}} & \textcolor{blue}{\textbf{1.43 ($\pm$0.95)}}& \textcolor{blue}{\textbf{0.78 ($\pm$0.25)}} & -- & \textcolor{blue}{\textbf{0.95  ($\pm$0.33)}}   \\
PAN (ResNet34)\cite{li2018pyramid}   & 0.86 ($\pm$0.20)  & 2.29 ($\pm$1.87) & 1.10 ($\pm$0.47) & -- & 1.42 ($\pm$0.52) \\
PAN (ResNet101) \cite{li2018pyramid}   & 0.74 ($\pm$0.19)  & 1.79 ($\pm$1.35) & 0.96 ($\pm$0.48) & -- & 1.17 ($\pm$0.47) \\
DAF \cite{wang18d}  & 0.76 ($\pm$0.17) & 1.84 ($\pm$1.33) & 1.02 ($\pm$0.66)& -- &  1.21 ($\pm$0.46)  \\
UNet Attention \cite{schlemper2019attention}  & 0.69 ($\pm$0.18) & 1.58 ($\pm$1.12) &  0.79 ($\pm$0.29) & --&  1.02 ($\pm$0.40)  \\
Proposed (MS-Dual-Guided)  & \textcolor{red}{\textbf{0.58 ($\pm$0.14)}}  & \textcolor{red}{\textbf{1.40 ($\pm$1.02)}}&  \textcolor{red}{\textbf{0.71 ($\pm$0.31)}} & -- &  \textcolor{red}{\textbf{0.90 ($\pm$0.36)}} \\
\midrule
\midrule
\end{tabular}

\caption{Comparison of the proposed network to other state-of-the-art architectures on the BRATS 2018 dataset (multi-organ segmentation on MRI task). The values show the average result of the experiments averaged over the 3 folds. Best results are represented in red bold, while blue is used to highlight the second best performance.}
\label{table:sota_comp_supplemental_brats}
\end{table*}

\begin{table}[]
    \centering
    \begin{tabular}{l|ccccc}
    \toprule
 \multicolumn{5}{c}{\textbf{Model complexity}}\\
 \midrule
 \textbf{Model} &  \multicolumn{4}{c}{\# Params}\\
 & & 1 Iter & 2 Iter & 3 Iter & 5 Iter\\
  \midrule
  UNet & 31,030,853 & - & - &- & - \\
  PAN (ResNet34) & 21,323,991 & - & - &- & - \\
  PAN (ResNet101) & 42,675,415 & - & - &- & - \\
  UNet Attention & 34,877,681 & - & - &- & - \\
  DANet (ResNet101) & 68,475,961 & - & - & - & - \\
  Proposed(DAF) & 43,482,179 & - & - &- & - \\
  Proposed(PAM) & 43,486,343 & - & - &- & - \\
  Proposed(CAM) & 43,485,543 & - & - &- & - \\
  Proposed(DANet) & 43,980,179 & - & - &- & - \\
  MS-Dual (No guidance) & - & 43,485,831 & 44,411,103 & 45,337,675 & 47,190,819 \\
  MS-Dual-Guided & - & 50,531,399 & 58,499,679 &  66,470,539 & 82,412,259\\
  MS-Dual-Guided (No Deep Sup)& - & 50,530,099 & 58,498,379 & 66,467,939  & 82,407,059 \\
  MS-Dual-Guided (Dist) & - & 43,485,831 & 44,411,103 & 45,337,675  & 47,190,819\\

\bottomrule
    \end{tabular}
    \caption{Model complexity, measured in number of parameters, for the evaluated models.}
    \label{tab:complexity}
\end{table}
\end{document}